\def\dssdpreprint{}

\documentclass{article} 
\usepackage{iclr2027_conference,times}
\usepackage{graphicx}
\usepackage{booktabs,multirow,longtable}

\usepackage{amsmath,amsfonts,bm}

\def\eqref#1{equation~\ref{#1}}

\def\1{\bm{1}}

\def\vb{{\bm{b}}}
\def\vc{{\bm{c}}}

\def\ve{{\bm{e}}}

\def\vh{{\bm{h}}}

\def\vk{{\bm{k}}}

\def\vm{{\bm{m}}}

\def\vo{{\bm{o}}}

\def\vs{{\bm{s}}}

\def\vv{{\bm{v}}}

\def\vx{{\bm{x}}}
\def\vy{{\bm{y}}}
\def\vz{{\bm{z}}}

\def\mA{{\bm{A}}}
\def\mB{{\bm{B}}}
\def\mC{{\bm{C}}}

\def\mE{{\bm{E}}}
\def\mF{{\bm{F}}}
\def\mG{{\bm{G}}}
\def\mH{{\bm{H}}}
\def\mI{{\bm{I}}}
\def\mJ{{\bm{J}}}

\def\mM{{\bm{M}}}

\def\mQ{{\bm{Q}}}

\def\mS{{\bm{S}}}

\def\mU{{\bm{U}}}

\def\mW{{\bm{W}}}

\def\mSigma{{\bm{\Sigma}}}

\DeclareMathAlphabet{\mathsfit}{\encodingdefault}{\sfdefault}{m}{sl}
\SetMathAlphabet{\mathsfit}{bold}{\encodingdefault}{\sfdefault}{bx}{n}

\newcommand{\E}{\mathbb{E}}

\newcommand{\R}{\mathbb{R}}

\newcommand{\Cov}{\mathrm{Cov}}

\usepackage{amsthm}
\usepackage{algorithm,algpseudocode}
\newtheorem{proposition}{Proposition}
\newtheorem{corollary}{Corollary}
\usepackage[framemethod=tikz]{mdframed}
\definecolor{propositionborder}{HTML}{315B80}
\definecolor{propositionbackground}{HTML}{F0F6FC}
\mdfsetup{skipbelow=2pt}
\newenvironment{methodproposition}[1][]{%
  \begin{mdframed}[linewidth=0.7pt,linecolor=propositionborder,
    backgroundcolor=propositionbackground,roundcorner=5pt,
    innerleftmargin=8pt,innerrightmargin=8pt,
    innertopmargin=6pt,innerbottommargin=6pt,nobreak=true,
    skipabove=6pt]
  \begin{proposition}[#1]%
}{%
  \end{proposition}
  \end{mdframed}\vspace{-\parskip}%
}

\usepackage{hyperref}
\usepackage{url}

\newcommand{\titlealign}{}
\ifdefined\dssdpreprint
  \renewcommand{\titlealign}{\centering}
\fi
\title{\titlealign Dynamical State-Space Diffusion for \\ Spatiotemporal Prediction}

\newcommand{\authorrow}[1]{\makebox[\dimexpr\textwidth-2\tabcolsep\relax][c]{#1}}
\author{%
\authorrow{Woojun Lee$^{*}$ \qquad Hankyeol Kim$^{*}$ \qquad Pilsung Kang$^{\dagger}$}\\[4pt]%
\authorrow{\normalsize Department of Industrial Engineering, Seoul National University}\\%
\authorrow{\normalsize Seoul, Republic of Korea}\\[2pt]%
\authorrow{\small\ttfamily \{woojun\_lee, hankyeol, pilsung\_kang\}@snu.ac.kr}\\[4pt]%
\authorrow{\footnotesize $^{*}$Equal contribution. \qquad $^{\dagger}$Corresponding author.}%
}

\ifdefined\dssdpreprint
  \iclrfinalcopy
  \hypersetup{%
    pdftitle={Dynamical State-Space Diffusion for Spatiotemporal Prediction},
    pdfauthor={Woojun Lee, Hankyeol Kim, Pilsung Kang}}
\fi

\begin{document}

\maketitle
\ifdefined\dssdpreprint
  \lhead{Preprint.}
\fi

\begin{abstract}
Diffusion models for temporal prediction typically corrupt each future frame with independent noise, leaving temporal dependence entirely to the denoiser. We propose Dynamical State-Space Diffusion (DSSD), which learns the temporal structure of the corruption process itself. DSSD builds a temporal kernel from the damped oscillatory modes of a stable diagonal state-space model, whose learned decay rates and frequencies let the corruption capture multiple timescales and oscillatory correlations, while the denoiser architecture remains unchanged. Running this model on a finite clip, however, requires the state before the first frame, which is unobserved; starting from a zero state leaves the early frames with less noise than the diffusion schedule prescribes. We avoid this by using the Cholesky factor of the model's stationary covariance as the temporal operator. We prove that this operator is causal, exactly invertible, keeps each frame's noise variance equal to that of standard diffusion, and reduces to the identity at the clean end of the process, and that it is exactly the Kalman innovations form of the state-space model. The denoiser is trained to predict the temporally correlated noise, and sampling removes and reapplies the temporal mixing at each step. On scientific-field and robot-video datasets, DSSD outperforms diffusion baselines that share its denoiser, and the evolution of its learned modes during training shows that the temporal operator adapts to each dataset.
\end{abstract}

\section{Introduction}
\label{sec:introduction}

Diffusion models generate data by learning to reverse a forward process that gradually corrupts clean samples with Gaussian noise. This denoising view has led to strong image and video generators \citep{ho2020ddpm,song2021ddim,rombach2022ldm,ho2022video,blattmann2023svd}, and it has also become attractive for temporal prediction \citep{voleti2022mcvd,gao2023prediff,rasul2021timegrad,tashiro2021csdi}. In forecasting, a model must represent not one future but a distribution over plausible futures conditioned on an observed history, making diffusion a natural fit for uncertain spatiotemporal signals such as video, weather, and multivariate time series.

Most diffusion frameworks for temporal prediction use a forward process that adds temporally independent noise to each future frame, relying on the denoising network to learn temporal dependencies \citep{voleti2022mcvd,gao2023prediff}. However, neighboring states are often temporally coupled through motion, transport, or oscillation. This motivates incorporating temporal structure into the forward process itself, exposing the denoiser to temporally structured signal and noise.

Dynamical Diffusion (DyDiff) \citep{guo2025dydiff} incorporates temporal dependence into diffusion by applying an exponential moving average (EMA) operator to both the clean sequence and the diffusion noise. This introduces temporal correlations while preserving a closed-form Gaussian forward process and an invertible temporal operator for sampling. However, at each diffusion step, the EMA operator imposes a single exponential decay across frames. Its decay rate follows a predefined schedule rather than being learned from data, limiting the operator to one memory scale and preventing it from expressing oscillatory correlations.

We propose \emph{Dynamical State-Space Diffusion} (DSSD), a novel diffusion algorithm that incorporates a learnable state-space operator into forward corruption and reverse sampling. Inspired by S4D \citep{gu2022s4d}, the model combines multiple damped oscillatory modes whose decay rates, frequencies, and mixture weights are learned jointly with the denoiser. As illustrated in Figure 1, DSSD uses these learned dynamics to causally mix observed and future latents and to introduce temporal correlations into the diffusion noise. The state-space parameters are shared across sequences and spatial locations, while the denoiser architecture remains unchanged.

\begin{figure}[t]
\centering
\includegraphics[width=\textwidth]{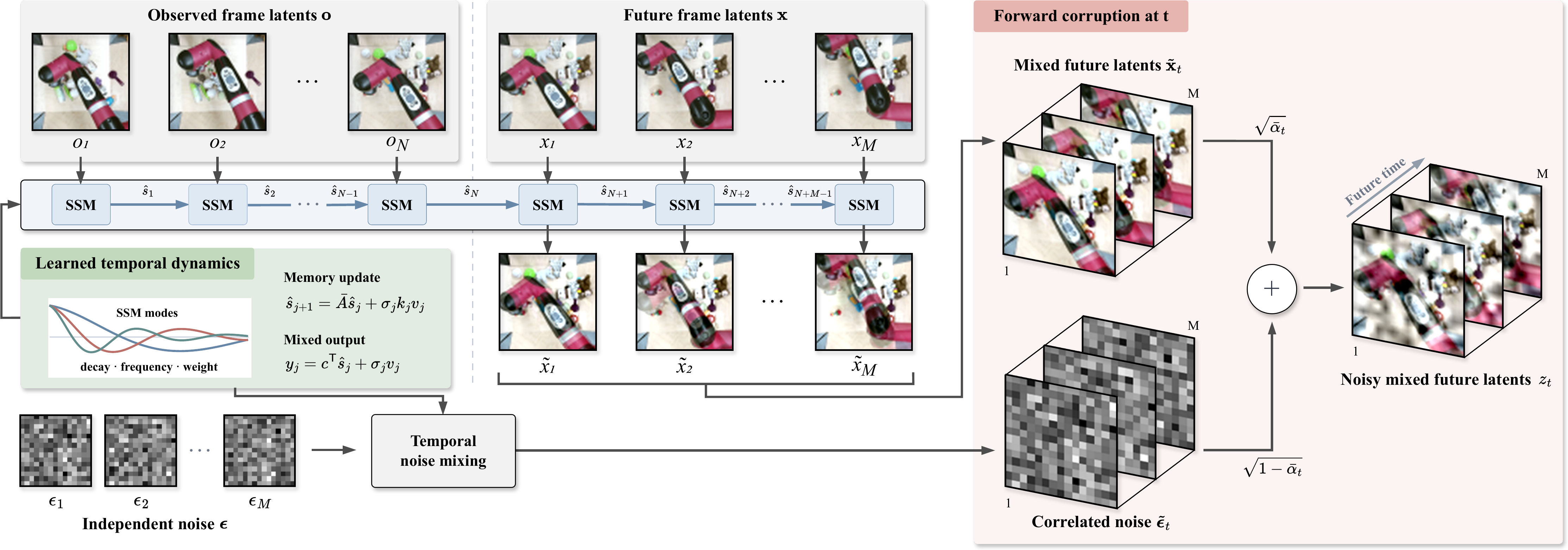}
 \caption{\textbf{DSSD forward process.} Learned state-space dynamics causally mix observed and future latents and separately correlate noise within the future window. The mixed future latents and correlated noise are combined at diffusion step \(t\) to form \(\mathbf{z}_t\).}
\label{fig:intro-comparison}
\end{figure}

Constructing a finite-window temporal operator from this learned kernel requires care at the start of the prediction window. Each output depends on both current and past noise inputs, including those preceding the window. Initializing the state-space model at zero omits the contribution of these earlier inputs, leaving the early frames with less noise variance than prescribed. This violates the requirement that every future frame receive the same noise variance at a given diffusion step. DSSD avoids this problem by using the Cholesky factor of the kernel's stationary finite-window covariance, yielding a causal and exactly invertible operator with unit marginal noise variance. We use corresponding factors to transform the signal and noise, train the denoiser to predict the resulting correlated noise, and adapt each sampling step to undo and reapply the temporal transform.

Our contributions are summarized as follows:

\begin{itemize}
\item We introduce DSSD, a dynamics-aware diffusion framework that uses a learnable diagonal state-space operator to define the temporal coordinates of the forward and reverse processes, rather than placing all temporal modeling burden on the denoiser.
\item We construct a stationary-covariance Cholesky operator that avoids finite-window boundary effects, is causal and exactly invertible, and preserves unit marginal noise variance. This is equivalent to a Kalman innovations realization, with DyDiff's EMA operator recovered as a single-mode case.
\item Experiments on scientific-field (SEVIR, Turbulence) and robot-video (BAIR, RoboNet) datasets show that DSSD outperforms diffusion baselines sharing the same denoiser. Analysis of mode evolution during training shows that DSSD's temporal operator adapts to each dataset.

\end{itemize}
    
\section{Related Work}
\label{sec:related-work}

\paragraph{Diffusion models for temporal prediction.}
Conditional diffusion models have been applied to video prediction \citep{voleti2022mcvd,ho2022video}, precipitation nowcasting \citep{gao2023prediff}, and probabilistic time-series forecasting \citep{rasul2021timegrad,tashiro2021csdi,alcaraz2023sssd}. These methods usually model temporal dependence through the conditioning scheme and the denoiser, for example by masking past frames, generating autoregressively, or adding temporal attention and convolution layers. The forward process remains standard, where each future time step receives independent noise. DSSD instead changes the temporal geometry of the forward process and leaves the denoiser unchanged, so it applies to existing diffusion algorithms.

\paragraph{Structured and learned forward processes.} For video generation, temporally correlated noise priors share noise components across frames \citep{ge2023pyoco} or warp noise along optical flow \citep{chang2024warped}; these methods structure the noise but do not apply a matched temporal transform to the clean signal. Blurring and inverse heat dissipation apply a fixed linear operator to the signal, but along the spatial frequency axis rather than the prediction horizon \citep{hoogeboom2023blurring,rissanen2023heat}. Other works learn part of the forward process, such as the noise schedule \citep{kingma2021vdm} or a data-dependent transformation of the corruption \citep{bartosh2024nfdm}. DYffusion \citep{ruhlingcachay2023dyffusion} replaces diffusion steps with a learned temporal interpolator, so its reverse process advances physical time, whereas DSSD keeps Gaussian diffusion and changes only its covariance. DyDiff \citep{guo2025dydiff} applies a fixed EMA operator to both signal and noise, limiting its ability to represent multi-scale or oscillatory temporal correlations. DSSD learns this operator as a state-space model jointly with the denoiser, keeps the forward process causal, invertible and variance preserving, and recovers DyDiff as an exact special case.

\paragraph{State-space models.} Structured state-space models parameterize linear recurrences with stable, structured transition matrices and scale to long sequences \citep{gu2022s4,gu2022s4d}, and selective variants make these dynamics input dependent \citep{gu2023mamba}. In diffusion models, SSMs have mainly served as network layers, either as attention-free backbones \citep{yan2024diffussm} or as temporal layers of time-series denoisers \citep{alcaraz2023sssd}. Linear-Gaussian state-space models also underlie probabilistic forecasting, where they describe how the data are generated \citep{rangapuram2018deepstate}. In contrast, DSSD uses an SSM to define the temporal covariance of the forward process, whose stationary innovations factor gives the diffusion forward operator.
\section{Preliminaries}
\label{sec:preliminaries}

\subsection{Conditional diffusion for temporal prediction}
\label{sec:prelim-diffusion}

Given $N$ observed frames of a spatiotemporal signal, we predict the next $M$. Frames are
encoded by a pretrained autoencoder \citep{rombach2022ldm} into observed latents
$\vo\in\R^{N\times d}$ and future latents $\vx\in\R^{M\times d}$, where $d$ is the dimension
of a flattened latent frame. We model $p(\vx\mid\vo)$ by diffusing only the future while
conditioning the denoiser on the clean observations.

Diffusion models \citep{ho2020ddpm} corrupt $\vx$ with Gaussian noise. For a
variance-preserving schedule $\alpha_t\in(0,1)$, $t=1,\dots,T$, whose product
$\bar\alpha_t=\prod_{s=1}^{t}\alpha_s$ decreases from nearly $1$ to $0$,
\begin{equation}
\vz_t=\sqrt{\bar\alpha_t}\,\vx+\sqrt{1-\bar\alpha_t}\,\bm{\epsilon},
\qquad \bm{\epsilon}\sim\mathcal{N}(\bm{0},\mI).
\label{eq:forward}
\end{equation}
A denoiser $f_\theta$ is trained to predict the added noise by minimizing
$\E_{\vx,\vo,t,\bm{\epsilon}}\lVert f_\theta(\vz_t,\vo,t)-\bm{\epsilon}\rVert_2^2$,
with $t$ sampled uniformly. The added noise is independent across future frames, while the
denoising network learns temporal dependencies from the data.

Futures are generated from Gaussian noise by repeated denoising, for example with the deterministic DDIM
step \citep{song2021ddim}
$\vz_{t'}=\sqrt{\bar\alpha_{t'}}\,\hat{\vx}+\sqrt{1-\bar\alpha_{t'}}\,\hat{\bm{\epsilon}}$ for $t'<t$,
where $\hat{\bm{\epsilon}}=f_\theta(\vz_t,\vo,t)$ and
$\hat{\vx}=(\vz_t-\sqrt{1-\bar\alpha_t}\,\hat{\bm{\epsilon}})/\sqrt{\bar\alpha_t}$.
Throughout, temporal matrices act along the frame axis and apply the same transformation
independently to each latent coordinate.

\subsection{State-space models}
\label{sec:prelim-ssm}

Structured state-space models (SSMs) \citep{gu2022s4,gu2022s4d} describe a sequence through a
hidden state that summarizes its past. For the continuous-time linear system
$\dot{\vh}(\tau)=\mA\vh(\tau)+\vb\,e(\tau)$, $y(\tau)=\vc^{\top}\vh(\tau)$,
zero-order-hold discretization with step size $\Delta>0$ gives the recurrence
\begin{equation}
\vh_k=\bar{\mA}\vh_{k-1}+\bar{\vb}\,e_k,
\qquad y_k=\vc^{\top}\vh_k,
\qquad \bar{\mA}=e^{\Delta\mA},
\label{eq:ssm-recurrence}
\end{equation}
with $\bar{\vb}$ the discretized input vector. From a zero initial state, unrolling gives a causal convolution
$y_k=\sum_{\ell\ge0}\bar K_\ell\,e_{k-\ell}$ whose kernel $\bar K_\ell=\vc^{\top}\bar{\mA}^{\ell}\bar{\vb}$
weights the input $\ell$ steps in the past.

S4D \citep{gu2022s4d} makes $\mA$ diagonal. Taking $P$ eigenvalues together with their complex
conjugates, $\mA=\operatorname{diag}(\lambda_1,\dots,\lambda_P,\lambda_1^*,\dots,\lambda_P^*)$
with $\operatorname{Re}\lambda_n<0$, and conjugate coefficients for conjugate states, the kernel
becomes a real-valued sum of damped oscillatory modes,
\begin{equation}
\bar K_\ell=\sum_{n=1}^{P} r_n\,e^{\ell\Delta\operatorname{Re}\lambda_n}
\cos\!\big(\ell\Delta\operatorname{Im}\lambda_n+\psi_n\big),
\label{eq:s4d-kernel}
\end{equation}
where the amplitude $r_n\ge0$ and phase $\psi_n$ are determined by $\bar{\vb}$ and $\vc$. Each
mode is a damped oscillation with decay rate $-\operatorname{Re}\lambda_n$ and frequency
$\operatorname{Im}\lambda_n$: one zero-frequency mode gives a geometric decay, and several modes
combine decay rates and oscillations in one kernel.

When driven by i.i.d.\ zero-mean, unit-variance inputs from the infinite past, a stable SSM has a stationary
output whose covariance depends only on the lag $\delta\ge0$,
\begin{equation}
\Cov(y_k,y_{k+\delta})=\sum_{\ell\ge0}\bar K_\ell\,\bar K_{\ell+\delta}.
\label{eq:ssm-autocov}
\end{equation}

\section{Dynamical State-Space Diffusion}
\label{sec:method}

We first define how a temporal operator enters the forward process, then build it from a
learned SSM kernel through its stationary covariance and Cholesky factor, and finally
describe training, sampling and the scope of the guarantees.

\subsection{Dynamics-aware forward process}
\label{sec:method-forward}

Following \citet{guo2025dydiff}, we couple the strength of temporal mixing to the noise level
through a dynamics schedule $\bar\gamma_t=1-\eta\,(1-\bar\alpha_t)\in[1-\eta,1]$ with
$\eta\in[0,1]$, so that $\bar\gamma_t\to1$ (no mixing) as the data become clean and mixing is
strongest at high noise. Let $\mW^{(L)}_{\gamma}\in\R^{L\times L}$ denote a temporal operator on $L$ frames
at dynamics level $\gamma\in(0,1]$, and partition the operator on the full window as
\begin{equation}
\mW^{(N+M)}_{\gamma}=\begin{bmatrix}\mW^{oo}_{\gamma}&\bm{0}\\
\mW^{fo}_{\gamma}&\mW^{ff}_{\gamma}\end{bmatrix}.
\label{eq:block}
\end{equation}
The blocks act on the $N$ observed and $M$ future frames of
Section~\ref{sec:prelim-diffusion}: $\mW^{fo}_\gamma$ carries the observed history into the
future window, $\mW^{ff}_\gamma$ mixes future frames with one another, and the zero block keeps
the observations free of any influence from the future. DSSD replaces \eqref{eq:forward} by
\begin{equation}
\vz_t=\sqrt{\bar\alpha_t}\,\big(\mW^{fo}_{\bar\gamma_t}\vo+\mW^{ff}_{\bar\gamma_t}\vx\big)
+\sqrt{1-\bar\alpha_t}\,\mW^{(M)}_{\bar\gamma_t}\bm{\epsilon},
\qquad \bm{\epsilon}\sim\mathcal{N}(\bm{0},\mI).
\label{eq:dssd-forward}
\end{equation}
The clean future is mixed together with the observed history, so the first future frame
inherits the temporal state of the observations, whereas the noise is mixed within the future
window alone, without an observed prefix. For fixed dynamics parameters, $q(\vz_t\mid\vo,\vx)$ is Gaussian
with mean $\sqrt{\bar\alpha_t}(\mW^{fo}\vo+\mW^{ff}\vx)$ and covariance
$(1-\bar\alpha_t)\,\mW^{(M)}\mW^{(M)\top}$, so training inputs are still sampled in closed form,
and $\mW=\mI$ recovers \eqref{eq:forward}. As in DDIM \citep{song2021ddim}, the process is
specified by these marginals and a coupling in which all diffusion steps share the same
$(\vx,\bm{\epsilon})$; a Markov forward chain is not required.

We require the operator to be
(i)~\emph{causal}, mixing a frame only with itself and preceding frames, i.e., lower triangular;
(ii)~\emph{invertible}, so that clean frames and noise can be recovered during sampling;
(iii)~of \emph{unit per-frame variance},
$\operatorname{diag}(\mW\mW^\top)=\bm{1}$, so that every frame receives noise of variance
$1-\bar\alpha_t$ exactly as in \eqref{eq:forward} and the single noise level $t$ given to the
denoiser describes every frame; and (iv)~the \emph{identity} at $\gamma=1$, so that the
clean end of the process is unchanged.
These are design requirements for DSSD, rather than necessary conditions for diffusion models in general.

\subsection{State-space temporal kernel}
\label{sec:method-kernel}

\begin{figure}[t]
\centering
\includegraphics[width=\textwidth]{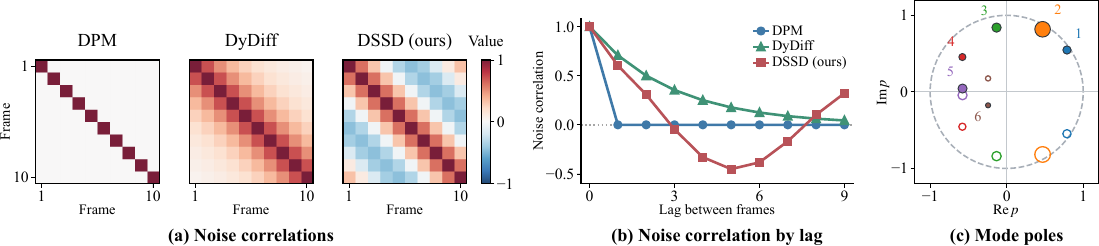}
\caption{\textbf{Noise correlations and state-space modes in DSSD.}
(a) Illustrative noise correlation matrices for DPM, DyDiff, and DSSD.
(b) Corresponding noise correlations by lag: DyDiff's EMA decays exponentially,
whereas DSSD can express oscillatory correlations.
(c) An illustrative six-mode DSSD configuration shows discrete poles
$p_n=e^{\Delta_\gamma\lambda_n}$ and their conjugates inside the unit circle.
Filled and hollow markers denote conjugate partners.}
\label{fig:temporal-correlations}
\end{figure}

DSSD describes the temporal correlation of the mixed frames with a diagonal SSM
(Section~\ref{sec:prelim-ssm}) whose discretization step is set by the dynamics level. For $P$
modes with learnable decay rates $a_n>0$ and frequencies $\omega_n\ge0$, we set
\begin{equation}
\lambda_n=-\tfrac{a_n}{2}+i\,\omega_n,\qquad \Delta_\gamma=-\log(1-\gamma).
\label{eq:dssd-poles}
\end{equation}
For $0<\gamma<1$, each discrete pole $p_n=e^{\Delta_\gamma\lambda_n}$ has modulus $(1-\gamma)^{a_n/2}<1$.
Figure~\ref{fig:temporal-correlations}(c) illustrates these stable poles and their conjugate pairs. As
$\gamma\to1$ the step $\Delta_\gamma$ grows without bound and every mode forgets immediately,
whereas at high noise the step is smallest and memory is longest. The step is determined by
the dynamics schedule, while the learned rates and frequencies control individual modes;
$a_n=1$ reproduces the EMA decay $\sqrt{1-\gamma}$ of DyDiff. We use amplitudes
$r=\operatorname{softmax}(\rho)$ and set
every phase $\psi_n$ in \eqref{eq:s4d-kernel} to zero, giving positive real modal coefficients,
and normalize the resulting kernel to unit $\ell_2$ norm,
\begin{equation}
\tilde u_\ell=\sum_{n=1}^{P}r_n\,e^{\ell\Delta_\gamma\operatorname{Re}\lambda_n}
\cos\!\big(\ell\Delta_\gamma\operatorname{Im}\lambda_n\big),
\qquad
u_\ell=\tilde u_\ell\Big/\Big(\textstyle\sum_{\ell'\ge0}\tilde u_{\ell'}^{\,2}\Big)^{1/2}.
\label{eq:dssd-kernel}
\end{equation}
This zero-phase choice fixes $\tilde u_0=\sum_nr_n=1$, which keeps the normalization
bounded away from zero. We use this kernel for memory of past inputs and add a separate
current-input term. The resulting impulse response is
\begin{equation}
w_0=\sqrt{\gamma},\qquad w_k=\sqrt{1-\gamma}\;u_{k-1}\quad(k\ge1),
\qquad \textstyle\sum_{k\ge0}w_k^2=1 .
\label{eq:dssd-impulse}
\end{equation}
Thus the SSM kernel is delayed by one frame: $w_0$ weights the current input and $w_k$
weights an input $k$ frames in the past. For independent unit-variance inputs, $\gamma$
is the variance contributed by the current input, while the kernel distributes the remaining
variance $1-\gamma$ over the past. At $\gamma=1$, we define $w=(1,0,\dots)$ by its limit.
The dynamics parameters $\phi=\{\log a_n,\omega_n,\rho_n\}_{n=1}^{P}$ comprise only $3P$
scalars, shared across frames, pixels, samples and diffusion steps.
Learning these parameters jointly with the denoiser allows the noise correlations to
adapt to the data's temporal structure beyond a fixed exponential decay.

\subsection{Causal innovations operator}
\label{sec:method-operator}

Applying the impulse response \eqref{eq:dssd-impulse} as a convolution over a finite window
requires inputs preceding that window, or their accumulated state. Starting from a zero state discards the memory of
earlier inputs, so the early frames have variance below one and (iii) fails: the first frame
has variance $w_0^2=\gamma$. For example, at $\gamma=0.5$ it receives only half of the
prescribed noise variance. DSSD instead assumes that the SSM has reached its stationary regime. By
\eqref{eq:ssm-autocov}, driving the impulse response \eqref{eq:dssd-impulse} with i.i.d.\
unit-variance inputs yields a stationary unit-variance process, and DSSD uses the causal factor
of the covariance of $L$ consecutive outputs,
\begin{equation}
\kappa_\delta(\gamma)=\sum_{k\ge0}w_k\,w_{k+\delta},
\qquad
\big[\mJ^{(L)}_{\gamma}\big]_{ij}=\kappa_{|i-j|}(\gamma),
\qquad
\mW^{(L)}_{\gamma}=\operatorname{chol}\big(\mJ^{(L)}_{\gamma}\big),
\label{eq:dssd-operator}
\end{equation}
where $\operatorname{chol}$ returns the unique lower-triangular factor with positive diagonal
satisfying $\mW\mW^\top=\mJ$, and all infinite sums admit closed forms
(Appendix~\ref{app:closed-form}). For Gaussian inputs, row $j$ of $\mW$ expresses output $j$
as the part predictable from preceding outputs plus an independent innovation, whose scale
is $[\mW]_{jj}$. Thus $u$ defines the memory kernel, $w$ includes the current-input term,
$\mJ$ records stationary correlations, and $\mW$ realizes them on a finite window.
Figure~\ref{fig:temporal-correlations}(a--b) compares these noise correlations with
those of standard diffusion and DyDiff.

\begin{methodproposition}[Admissibility]
\label{prop:admissible}
For every $\gamma\in(0,1]$, window length $L$ and parameters $\phi$:
(a)~$\mJ^{(L)}_\gamma$ is positive definite, so $\mW^{(L)}_\gamma$ exists and is causal and
invertible;
(b)~$\operatorname{diag}\big(\mW^{(L)}_\gamma\mW^{(L)\top}_\gamma\big)=\bm{1}$;
(c)~$[\mW^{(L)}_\gamma]_{jj}\ge\sqrt{\gamma}$ for all $j$, hence
$[\mW^{(L)}_{\bar\gamma_t}]_{jj}\ge\sqrt{1-\eta}$ at every diffusion step;
(d)~the leading $N\times N$ block of $\mW^{(N+M)}_\gamma$ equals $\mW^{(N)}_\gamma$, and
$\mW^{fo}_\gamma\mW^{fo\top}_\gamma+\mW^{ff}_\gamma\mW^{ff\top}_\gamma
=\mW^{(M)}_\gamma\mW^{(M)\top}_\gamma$; and
(e)~$\mW^{(L)}_\gamma\to\mW^{(L)}_1=\mI$ as $\gamma\to1$.
\end{methodproposition}

Properties (a), (b) and (e) establish requirements (i)--(iv), and (c) bounds from below every
divisor in the triangular solves used during sampling. The schedule-wide bound is strictly
positive when $\eta<1$; for $\eta=1$, invertibility requires $\bar\alpha_t>0$.
These bounds do not guarantee a well-conditioned inverse, which we check numerically
(Appendix~\ref{app:practical}). The
first part of (d) makes the operator consistent with the observed history, so the signal
branch transforms $\vo$ and continues into the future; the second part is used in
Section~\ref{sec:method-analysis}. All proofs are given in Appendix~\ref{app:proofs}.

The Cholesky factor also has a state-space realization. It is an innovations representation
of the stationary process defined by $w$, rather than the original S4D convolution matrix.

\begin{methodproposition}[State-space recurrence]
\label{prop:recurrence}
For $\gamma\in(0,1)$, $\mW^{(L)}_\gamma$ is the finite-window Kalman innovations realization
\citep{kailath2000linear} of the SSM of Section~\ref{sec:method-kernel}, started at its
stationary state covariance and zero prior state mean: $\vy=\mW^{(L)}_\gamma\ve$ is computed by a linear recurrence with a
$2P$-dimensional state (Appendix~\ref{app:recurrence}) whose innovation scales satisfy
$\sigma_j=[\mW^{(L)}_\gamma]_{jj}$, $\sigma_0=1$ and $\sigma_{j+1}\le\sigma_j$.
\end{methodproposition}

The state is the best estimate of the SSM state given the frames read so far, and the
innovation scale shrinks as the state becomes better determined. In this view the two branches
of \eqref{eq:dssd-forward} use different histories: the signal branch runs the
recurrence with the observed latents as inputs and continues with the future latents, whereas the noise branch
restarts from the stationary initialization. For long windows, this recurrence reduces the
cost of applying the operator to linear in $L$ (Appendix~\ref{app:recurrence}).

\begin{methodproposition}[EMA and richer correlations]
\label{prop:richer}
Let $\gamma\in(0,1)$. With a single mode, $a_1=1$ and $\omega_1=0$, $\mW^{(L)}_\gamma$ is the EMA
operator of DyDiff \citep{guo2025dydiff}, $y_0=e_0$ and
$y_j=\sqrt{1-\gamma}\,y_{j-1}+\sqrt{\gamma}\,e_j$, whose autocorrelation
$\kappa_\delta(\gamma)=(1-\gamma)^{\delta/2}$ is positive at every lag $\delta$. In contrast,
the DSSD family admits a single oscillating mode with $\kappa_1(\gamma)<0$.
\end{methodproposition}

At each fixed dynamics level, the unrestricted DSSD covariance family therefore includes
the EMA covariance (Corollary~\ref{cor:ema}) and correlations beyond it: an
oscillating mode can make neighboring frames anticorrelated, a pattern that a monotonically
decaying memory cannot produce (Appendix~\ref{app:richer}).

\subsection{Training and sampling}
\label{sec:method-training}

\paragraph{Training.}
The denoiser parameters $\theta$ and the dynamics parameters $\phi$ are trained jointly by
regressing the mixed noise,
\begin{equation}
\mathcal{L}(\theta,\phi)=\E_{\vx,\vo,t,\bm{\epsilon}}
\big\lVert f_\theta(\vz_t,\vo,t)-\mW^{(M)}_{\bar\gamma_t}\bm{\epsilon}\big\rVert_2^2,
\label{eq:dssd-loss}
\end{equation}
with $\vz_t$ given by \eqref{eq:dssd-forward} and $t$ sampled uniformly. For a fixed operator,
the minimizer is the posterior mean of the mixed noise. Because the noise enters $\vz_t$ only
through $\mW^{(M)}_{\bar\gamma_t}\bm{\epsilon}$, the score of $q(\vz_t\mid\vo)$ is a fixed linear
transform of this posterior mean (Appendix~\ref{app:score}), so the correlated target requires
no whitening. Since every coordinate of the target has unit variance for every $\phi$,
learning $\phi$ cannot lower the loss by shrinking the noise. Because $\phi$ enters both the corrupted input and the target, learning $\phi$ selects the
temporal coordinates in which the denoiser operates rather than identifying the physical
dynamics of the data. Initialization and numerical checks are given in
Appendices~\ref{app:zero-frequency} and~\ref{app:practical}.
Algorithm~\ref{alg:dssd-training} summarizes joint training of the denoiser and dynamics parameters.

\paragraph{Sampling.}
We use deterministic DDIM steps that undo the mixing of the current step and reapply that of
the next one. Given the prediction $\hat{\bm{\epsilon}}_c=f_\theta(\vz_t,\vo,t)$ at step $t$ and
the next step $t'<t$,
\begin{align}
\hat{\vx}&=\big(\mW^{ff}_{\bar\gamma_t}\big)^{-1}
\Big(\tfrac{1}{\sqrt{\bar\alpha_t}}\big(\vz_t-\sqrt{1-\bar\alpha_t}\,\hat{\bm{\epsilon}}_c\big)
-\mW^{fo}_{\bar\gamma_t}\vo\Big),
\qquad
\hat{\bm{\epsilon}}=\big(\mW^{(M)}_{\bar\gamma_t}\big)^{-1}\hat{\bm{\epsilon}}_c,
\nonumber\\
\vz_{t'}&=\sqrt{\bar\alpha_{t'}}\,\big(\mW^{fo}_{\bar\gamma_{t'}}\vo+\mW^{ff}_{\bar\gamma_{t'}}\hat{\vx}\big)
+\sqrt{1-\bar\alpha_{t'}}\,\mW^{(M)}_{\bar\gamma_{t'}}\hat{\bm{\epsilon}}.
\label{eq:dssd-ddim}
\end{align}
Sampling starts from $\vz_T=\mW^{(M)}_{\bar\gamma_T}\bm{\xi}$ with
$\bm{\xi}\sim\mathcal{N}(\bm{0},\mI)$, and the last step uses $\bar\alpha=\bar\gamma=1$, where
the operator is the identity and $\hat{\vx}$ is returned. Both inverses are triangular solves
whose positive divisors satisfy Proposition~\ref{prop:admissible}(c), and with $\mW=\mI$
the update reduces to the standard DDIM step. Algorithm~\ref{alg:dssd-sampling} gives the
complete sampling procedure, including initialization and the clean endpoint.
The following result justifies this update.

\begin{methodproposition}[Sampler consistency]
\label{prop:sampler}
Fix the dynamics parameters $\phi$, assume finite second moments of the data, and let
$f_\theta$ be the unrestricted minimizer of \eqref{eq:dssd-loss}. For $0<\bar\alpha_t<1$,
$\hat{\vx}=\E[\vx\mid\vz_t,\vo]$, $\hat{\bm{\epsilon}}=\E[\bm{\epsilon}\mid\vz_t,\vo]$, and
\eqref{eq:dssd-ddim} returns $\E[\vz_{t'}\mid\vz_t,\vo]$ under the shared $(\vx,\bm{\epsilon})$
coupling. On an interior time interval where $\bar\alpha$ is twice continuously
differentiable and $0<\bar\alpha<1$, $\vz_{t'}=\vz_t+(t'-t)\,\vv_t(\vz_t)+O\big((t'-t)^2\big)$ for a velocity
field $\vv_t$ whose continuity equation is solved by the marginals $q(\vz_t\mid\vo)$.
\end{methodproposition}

Under the flow regularity conditions in Appendix~\ref{app:sampler},
\eqref{eq:dssd-ddim} is a first-order discretization of a flow that transports the
marginals of the forward process \citep{albergo2025interpolants}, as DDIM is for standard
diffusion. With the ideal denoiser, its last step returns the posterior mean of the future.

\subsection{What the temporal operator changes}
\label{sec:method-analysis}

Since every operator is invertible, one may ask whether DSSD merely reparameterizes standard
diffusion. It does not: the mixing varies with the diffusion step and differs between the
signal and noise branches of \eqref{eq:dssd-forward}. Let
$\mF_t=\big(\mW^{(M)}_{\bar\gamma_t}\big)^{-1}\mW^{ff}_{\bar\gamma_t}$, and let
$\vz_t^{\mathrm{std}}=\sqrt{\bar\alpha_t}\,\vx+\sqrt{1-\bar\alpha_t}\,\bm{\epsilon}$ be the standard
corruption \eqref{eq:forward} at the same noise level.

\begin{methodproposition}[Information ordering]
\label{prop:information}
For every $t$, $\mF_t\mF_t^\top\preceq\mI$. Consequently, given $\vo$, a variable with the same
joint distribution with $\vx$ as $\vz_t$ can be generated from $\vz_t^{\mathrm{std}}$ and
independent noise without access to $\vx$, and, for data with finite second moments,
\begin{equation}
\min_g\,\E\big\lVert\vx-g(\vz_t^{\mathrm{std}},\vo)\big\rVert^2
\;\le\;\min_g\,\E\big\lVert\vx-g(\vz_t,\vo)\big\rVert^2 .
\label{eq:information}
\end{equation}
\end{methodproposition}

Temporal mixing therefore creates no information about the future at any single noise level; for
the EMA operator, $\mF_t=\operatorname{diag}(\sqrt{\bar\gamma_t},1,\dots,1)$ attenuates only the
direction of the first future frame (Appendix~\ref{app:information}). DSSD instead changes
the geometry of the regression problem that a finite denoiser solves across noise levels,
and the path along which samples are generated. These changes may help a finite denoiser and
sampler, but the proposition does not establish such a benefit; whether they help is tested
empirically in Section~\ref{sec:experiments}.

\section{Experiments}
\label{sec:experiments}

Proposition~\ref{prop:information} rules out an information advantage from temporal mixing
at a fixed noise level. We therefore test whether learning the temporal operator improves
finite-step forecasting over standard diffusion DPM and DyDiff. We evaluate scientific fields from SEVIR
\citep{veillette2020sevir} and Turbulence \citep{wang2020turbulence}, and robot videos from BAIR \citep{ebert2017visual} and RoboNet \citep{dasari2019robonet}.

\subsection{Experimental setup}
\label{sec:experiments-setup}
Within each dataset, all methods use the same frozen framewise VAE and Stable Video
Diffusion-inspired video U-Net \citep{blattmann2023svd}. We train for
50{,}000 optimizer updates and use a shared 50-step DDIM sampling grid
\citep{song2021ddim}.
Task windows and training settings are detailed in Appendix~\ref{app:experiments}, and
example forecasts for all four datasets in Appendix~\ref{app:additional-results-qualitative}.
Metrics include the continuous ranked probability score (CRPS), critical success index (CSI)
and Fr\'echet video distance (FVD); all are defined in Appendix~\ref{app:experiments-metrics}.
In Tables~\ref{tab:scientific-results} and~\ref{tab:robot-results}, bold and underline mark the best and second-best scores for each dataset and metric.

\subsection{Scientific-field prediction}
\begin{table}[t]
\centering
\footnotesize
\setlength{\tabcolsep}{3pt}
\caption{Scientific-field prediction results.}
\label{tab:scientific-results}
\begin{tabular*}{\linewidth}{@{\extracolsep{\fill}}llccccccc@{}}
\toprule
\multirow{2}{*}{Dataset} & \multirow{2}{*}{Method} &
\multicolumn{3}{c}{CRPS $\downarrow$} & \multirow{2}{*}{CSI w5 $\uparrow$} &
\multirow{2}{*}{\shortstack{Temporal\\variogram $\downarrow$}} &
\multirow{2}{*}{\shortstack{Energy\\score $\downarrow$}} &
\multirow{2}{*}{RMSE $\downarrow$} \\
\cmidrule(lr){3-5}
& & Raw & $\mathrm{w8}_{\mathrm{avg}}$ & $\mathrm{w8}_{\mathrm{max}}$ & & & & \\
\midrule
\multirow{3}{*}{SEVIR}
  & DPM         & \underline{4.174} & \underline{2.737} & \underline{6.599} & \underline{0.5518} & \underline{2.339} & \underline{10.862} & \underline{15.420} \\
  & DyDiff      & 4.642 & 3.169 & 7.955 & 0.3911 & 4.492 & 13.024 & 25.540 \\
  & DSSD (Ours) & \textbf{3.975} & \textbf{2.421} & \textbf{5.995} & \textbf{0.5707} & \textbf{2.129} & \textbf{10.263} & \textbf{14.585} \\
\midrule
\multirow{3}{*}{Turbulence}
  & DPM         & 0.01148 & \underline{0.00751} & \underline{0.01065} & 0.9641 & 0.00262 & 0.01920 & 0.02600 \\
  & DyDiff      & \underline{0.01100} & 0.00752 & 0.01123 & \underline{0.9674} & \underline{0.00260} & \underline{0.01791} & \underline{0.02427} \\
  & DSSD (Ours) & \textbf{0.00756} & \textbf{0.00440} & \textbf{0.00757} & \textbf{0.9779} & \textbf{0.00147} & \textbf{0.01178} & \textbf{0.01582} \\
\bottomrule
\end{tabular*}
\end{table}

\label{sec:experiments-scientific}
DSSD has the lowest scores on all
three CRPS variants, temporal variogram, energy score, and RMSE, and the highest
CSI on both datasets. Its raw CRPS is 3.975 versus 4.174 for the strongest
SEVIR baseline (DPM), and 0.00756 versus 0.01100 for the strongest Turbulence
baseline (DyDiff).

\subsection{Robot-video prediction}
\label{sec:experiments-robot}
\begin{table}[t]
\centering
\footnotesize
\setlength{\tabcolsep}{2pt}
\caption{Robot-video prediction results.}
\label{tab:robot-results}
\begin{tabular*}{\linewidth}{@{\extracolsep{\fill}}ll*{8}{c}@{}}
\toprule
\multirow{2}{*}{Dataset} & \multirow{2}{*}{Method} &
\multicolumn{4}{c}{Forecast metrics} & \multicolumn{4}{c}{Video metrics} \\
\cmidrule(lr){3-6}\cmidrule(lr){7-10}
& & CRPS $\downarrow$ & RMSE $\downarrow$ &
Energy $\downarrow$ & Vario. $\downarrow$ &
PSNR $\uparrow$ & SSIM $\uparrow$ & LPIPS $\downarrow$ & FVD $\downarrow$ \\
\midrule
\multirow{3}{*}{BAIR}
  & DPM         & 17.788 & 42.907 & 30.176 & \underline{11.078} & 15.036 & 0.5938 & 0.2246 & \underline{1329.2} \\
  & DyDiff      & \underline{17.691} & \underline{42.815} & \underline{30.092} & 11.412 & \underline{15.149} & \underline{0.5999} & \underline{0.2220} & 1417.4 \\
  & DSSD (Ours) & \textbf{16.825} & \textbf{42.367} & \textbf{29.661} & \textbf{8.548} & \textbf{15.717} & \textbf{0.6205} & \textbf{0.2007} & \textbf{766.0} \\
\midrule
\multirow{3}{*}{RoboNet}
  & DPM         & 8.782 & \underline{23.523} & 15.130 & \underline{5.939} & \underline{22.680} & \textbf{0.7889} & \textbf{0.1370} & \underline{659.2} \\
  & DyDiff      & \underline{8.383} & 23.735 & \underline{14.705} & 6.504 & 22.630 & \underline{0.7865} & \underline{0.1377} & 770.6 \\
  & DSSD (Ours) & \textbf{8.108} & \textbf{23.485} & \textbf{14.548} & \textbf{4.409} & \textbf{22.693} & 0.7858 & 0.1379 & \textbf{430.2} \\
\bottomrule
\end{tabular*}
\end{table}
On BAIR, DSSD outperforms both DPM and DyDiff across all eight reported metrics.
On RoboNet, DSSD leads six of eight metrics, while DPM achieves the best SSIM and LPIPS.
On both datasets, DSSD shows substantial improvements in FVD and temporal variogram scores,
whereas RMSE gains are modest.

\subsection{Prediction quality across horizons}
\label{sec:experiments-horizons}

\begin{figure}[t]
\centering
\includegraphics[width=0.9\textwidth]{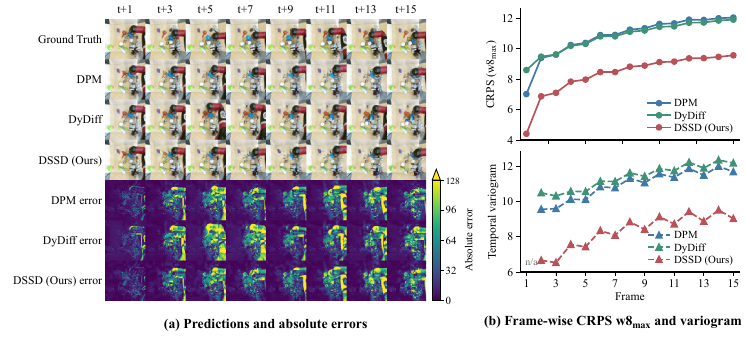}
\caption{BAIR forecast quality across future frames. (a) Example predictions and
absolute-error maps. (b) Test-set frame-wise pooled CRPS and temporal variogram
scores; lower is better.}
\label{fig:bair-horizons}
\end{figure}

Figure~\ref{fig:bair-horizons} compares prediction quality across horizons on BAIR.
Panel (a) shows an example predicted sequence and absolute-error maps for each method.
In this example, DSSD produces more accurate predictions throughout the sequence,
with particularly smaller absolute errors than DyDiff at intermediate time steps.
Panel (b) reports frame-wise CRPS($\mathrm{w8}_{\mathrm{max}}$) and temporal
variogram scores, averaged over all BAIR test windows at each prediction horizon.
DSSD maintains lower scores than both baselines throughout the evaluated horizons,
with its advantage persisting into the later frames. These results indicate
sustained improvements in probabilistic prediction and closer agreement with
ground-truth changes between consecutive frames.

\subsection{Analysis of learned temporal modes}
\label{sec:experiments-modes}
\begin{figure}[t]
\centering
\includegraphics[width=0.8\textwidth]{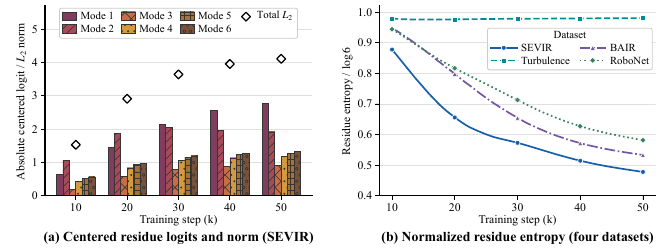}
\caption{DSSD mode-parameter diagnostics. (a) SEVIR centered residue logits and
their norm. (b) Normalized residue entropy on four datasets.}
\label{fig:mode-adaptation}
\end{figure}

The learned operator is compact, with only $3P=18$ parameters (Section~\ref{sec:method-kernel}).
Figure~\ref{fig:mode-adaptation} tracks these modes from 10{,}000 to 50{,}000
updates through the centered residue logits $|\rho_n-\bar\rho|$, their $L_2$ norm (the
residue-logit norm), and the normalized entropy of $r=\operatorname{softmax}(\rho)$
(Appendix~\ref{app:experiments-modes}).
The residue-logit norm measures logit spread around the mean, whereas lower entropy indicates more concentrated residue weights across modes.
On SEVIR, the residue-logit norm increases from 1.53 to 4.11 while entropy decreases.
Entropy also decreases on BAIR and RoboNet: SEVIR and RoboNet concentrate residue weight on the first mode, and BAIR on the second (Appendix~\ref{app:additional-results-modes}).
On Turbulence, the residue shares remain nearly uniform, yet the modes continue to evolve: at $\gamma=0.5$, the first mode's pole
radius decreases from $0.937$ to $0.889$, indicating faster decay and
shorter temporal memory. These observations indicate dataset-dependent adaptation of both mode weights and timescales.

\section{Conclusion}
\label{sec:conclusion}

We introduce DSSD, the novel diffusion framework for spatiotemporal prediction to make the coupling between spatial denoising and temporal dynamics jointly learnable. Its forward process adds noise while mixing frames through a learnable state-space operator, and its reverse process combines learned denoising with inversion of the temporal mixing. Joint training of the state-space dynamics and denoiser allows this coupled process to adapt to the data without changing the denoiser architecture. We establish a theoretical foundation through a stationary-covariance construction that ensures causal, invertible temporal mixing with the prescribed noise variance, together with a consistency analysis of the sampling update. Experiments on scientific fields and robot videos show improved probabilistic forecasts and temporal consistency, while analysis of the learned modes demonstrates dataset-dependent adaptation. We expect this framework to help accelerate the development of dynamics-aware video foundation models at higher resolutions and larger scales. Its learnable temporal dynamics may also support online adaptation to changing temporal patterns in streaming data, with potential applications in robotics.


%
%
%

\subsection*{AI use statement}

We used OpenAI Codex and Anthropic Claude to assist with implementing training, sampling,
evaluation, and experiment-management code based on our specifications. These tools also
assisted with typesetting and editing proofs, writing numerical verification scripts,
providing feedback on experimental design, drafting and editing the manuscript, searching
and summarizing literature, and formatting references.

The theoretical model, operator construction, mathematical claims, and proofs were developed
by the authors. AI assistance with proofs was limited to presentation and numerical checks;
it did not supply proof arguments. AI tools were not used to develop the conceptual framework,
propose or refine hypotheses, generate synthetic datasets, or conduct qualitative and
thematic data analysis. Translation was not applicable.

We reviewed all AI-assisted work, checked the proofs against our derivations, and verified
the algebra through independent numerical implementations. We tested the code and checked
citations against primary sources. The authors take responsibility for the final text,
claims, code, and other artifacts.

%
%

\subsection*{Ethics statement}

This work involves no human subjects, crowdsourced annotation, or personally identifiable information; no institutional review was required.

%
%

\subsection*{Reproducibility statement}

Section~\ref{sec:method} specifies the method and theoretical assumptions;
Appendix~\ref{app:proofs} provides complete proofs, and Appendix~\ref{app:practical} details
initialization and numerical implementation.
Appendix~\ref{app:experiments} specifies the data processing, model and optimization
settings, evaluation protocols, and compute resources used in our experiments.

We use publicly accessible datasets—SEVIR, BAIR, and RoboNet—and a simulation-based
benchmark, Turbulence. Third-party datasets are used in accordance with their terms and are
not redistributed.

We will release model weights, training and sampling code, and configurations for the models
and datasets used in this paper. The release will include environment specifications and
instructions for data preparation and reproducing the experiments. Run records and numerical
tests support verification of the experimental settings and theoretical identities.

\bibliography{iclr2027_conference}
\bibliographystyle{iclr2027_conference}

\appendix
\section{Mathematical Details and Proofs}
\label{app:proofs}

We first derive the stationary covariance, then prove the operator properties and its
state-space realization, and finally analyze regression and sampling. Throughout, the
dynamics parameters $\phi$ are fixed unless joint optimization is explicitly discussed;
$a_n>0$, $\omega_n\ge0$ and $r_n>0$ with $\sum_n r_n=1$.
For $\gamma\in(0,1)$, we omit $\gamma$ and $\phi$ when no confusion arises and write
$\nu=\sqrt{1-\gamma}$, $\Delta=\Delta_\gamma$,
$R_n=|e^{\Delta\lambda_n}|=\nu^{a_n}$ and $\vartheta_n=\Delta\omega_n$, so that
$\tilde u_\ell=\sum_{n}r_nR_n^\ell\cos(\ell\vartheta_n)$ in \eqref{eq:dssd-kernel}. As in the main
text, matrices act on the time axis and are shared by all coordinates of each frame.
The endpoint $\gamma=1$ is defined by $w=(1,0,\dots)$ and $\mW=\mI$; expressions involving
$\Delta$ are evaluated only for $\gamma<1$. We distinguish the SSM memory kernel $u_\ell$,
the full impulse response $w_k$, its stationary covariance $\mJ$, and the finite-window
innovations factor $\mW$.

\subsection{Notation}
\label{app:notation}

Table~\ref{tab:notation} collects the symbols used in Sections~\ref{sec:preliminaries}
and~\ref{sec:method} and in this appendix, grouped in the order in which they appear.
Bold lowercase letters denote vectors or stacked latent frames, bold uppercase letters denote
matrices, and all temporal matrices act along the frame axis, identically for every latent
coordinate. A few auxiliary symbols local to a single proof step, such as
$y^{\mathrm{pre}}_j$ in Appendix~\ref{app:admissible}, are defined where they appear.

{\small
\renewcommand{\arraystretch}{1.15}
\renewcommand{\eqref}[1]{Eq.~(\ref{#1})}
\begin{longtable}{@{}p{0.2\linewidth}p{0.58\linewidth}p{0.16\linewidth}@{}}
\caption{Notation.}\label{tab:notation}\\
\toprule
Symbol & Meaning & Defined in \\
\midrule
\endfirsthead
\toprule
Symbol & Meaning & Defined in \\
\midrule
\endhead
\bottomrule
\endfoot
\multicolumn{3}{@{}l}{\emph{Data and diffusion}}\\
$N$, $M$ & Numbers of observed and future frames & \S\ref{sec:prelim-diffusion} \\
$L$ & Length of the window on which a temporal operator acts; $L\in\{N,M,N+M\}$ & \S\ref{sec:method-forward} \\
$d$ & Dimension of a flattened latent frame & \S\ref{sec:prelim-diffusion} \\
$\vo\in\R^{N\times d}$ & Observed latents (clean conditioning history) & \S\ref{sec:prelim-diffusion} \\
$\vx\in\R^{M\times d}$ & Future latents to be predicted & \S\ref{sec:prelim-diffusion} \\
$T$, $t$ & Number of diffusion steps; current step $t\in\{1,\dots,T\}$ & \eqref{eq:forward} \\
$t'$ & Next, less noisy step of a sampling update, $t'<t$ & \S\ref{sec:prelim-diffusion} \\
$\alpha_t$, $\bar\alpha_t$ & Noise schedule and signal level $\bar\alpha_t=\prod_{s\le t}\alpha_s$ & \eqref{eq:forward} \\
$\bm{\epsilon}$ & White Gaussian noise, $\mathcal{N}(\bm{0},\mI)$, before temporal mixing & \eqref{eq:forward} \\
$\vz_t$ & Noisy future latents at step $t$ & \eqref{eq:forward}, \eqref{eq:dssd-forward} \\
$f_\theta$, $\theta$ & Denoiser and its parameters & \S\ref{sec:prelim-diffusion} \\
$p(\vx\mid\vo)$ & Conditional distribution of the future given the history & \S\ref{sec:prelim-diffusion} \\
$q(\vz_t\mid\vo)$ & Marginal of the forward process at step $t$ & \S\ref{sec:method-training} \\
\midrule
\multicolumn{3}{@{}l}{\emph{State-space models (generic)}}\\
$\tau$ & Continuous time & \S\ref{sec:prelim-ssm} \\
$\vh$, $e$, $y$ & State, scalar input and scalar output of an SSM & \eqref{eq:ssm-recurrence} \\
$\mA$, $\vb$, $\vc$ & State, input and output parameters & \S\ref{sec:prelim-ssm} \\
$\Delta$, $\bar{\mA}$, $\bar{\vb}$ & Step size and zero-order-hold discretization, $\bar{\mA}=e^{\Delta\mA}$ & \eqref{eq:ssm-recurrence} \\
$\bar K_\ell$ & Convolution kernel of the discretized SSM; weight of the input $\ell$ steps in the past & \eqref{eq:s4d-kernel} \\
$P$ & Number of complex-conjugate mode pairs & \S\ref{sec:prelim-ssm} \\
$\lambda_n$ & Continuous-time eigenvalue of mode $n$, $\operatorname{Re}\lambda_n<0$; in DSSD, $-a_n/2+i\,\omega_n$ & \S\ref{sec:prelim-ssm}, \eqref{eq:dssd-poles} \\
$r_n$, $\psi_n$ & Amplitude and phase of mode $n$ & \eqref{eq:s4d-kernel} \\
\midrule
\multicolumn{3}{@{}l}{\emph{Dynamics schedule and learned modes}}\\
$\eta$ & Mixing strength, $\eta\in[0,1]$ & \S\ref{sec:method-forward} \\
$\bar\gamma_t$ & Dynamics schedule, $1-\eta(1-\bar\alpha_t)\in[1-\eta,1]$; $1$ means no mixing & \S\ref{sec:method-forward} \\
$\gamma$ & Generic dynamics level, $\gamma\in(0,1]$ & \S\ref{sec:method-forward} \\
$a_n$, $\omega_n$ & Learned decay rate ($>0$) and frequency ($\ge0$) of mode $n$ & \eqref{eq:dssd-poles} \\
$\rho_n$ & Learned residue logit of mode $n$; $r=\operatorname{softmax}(\rho)$ & \S\ref{sec:method-kernel} \\
$\phi$ & Dynamics parameters $\{\log a_n,\omega_n,\rho_n\}_{n=1}^{P}$, $3P$ scalars & \S\ref{sec:method-kernel} \\
$\Delta_\gamma$ & Discretization step tied to the dynamics level, $-\log(1-\gamma)$ & \eqref{eq:dssd-poles} \\
$p_n$ & Discrete pole of mode $n$, $e^{\Delta_\gamma\lambda_n}$, with modulus $(1-\gamma)^{a_n/2}$ & \S\ref{sec:method-kernel} \\
\midrule
\multicolumn{3}{@{}l}{\emph{Kernel, covariance and operator}}\\
$\tilde u_\ell$, $u_\ell$ & Memory kernel before and after normalization to unit $\ell_2$ norm & \eqref{eq:dssd-kernel} \\
$w_k$ & Impulse response: $w_0=\sqrt\gamma$ weights the current input, $w_k=\sqrt{1-\gamma}\,u_{k-1}$ the input $k$ frames earlier & \eqref{eq:dssd-impulse} \\
$\kappa_\delta(\gamma)$ & Stationary autocorrelation at lag $\delta$, $\sum_kw_kw_{k+\delta}$; $\kappa_0=1$ & \eqref{eq:dssd-operator} \\
$\mJ^{(L)}_\gamma$ & Stationary covariance of $L$ consecutive outputs, $[\mJ]_{ij}=\kappa_{|i-j|}$ & \eqref{eq:dssd-operator} \\
$\operatorname{chol}$ & Lower-triangular Cholesky factor with positive diagonal & \eqref{eq:dssd-operator} \\
$\mW^{(L)}_\gamma$ & Temporal operator, $\operatorname{chol}(\mJ^{(L)}_\gamma)$ & \eqref{eq:dssd-operator} \\
$\mW^{oo}$, $\mW^{fo}$, $\mW^{ff}$ & Blocks of $\mW^{(N+M)}_\gamma$: history on history, history into future, future on future & \eqref{eq:block} \\
$\mW^{(M)}_\gamma$ & Operator applied to the noise, on the future window alone & \eqref{eq:dssd-forward} \\
\midrule
\multicolumn{3}{@{}l}{\emph{Innovations recurrence (Proposition~\ref{prop:recurrence})}}\\
$\ve$, $e_j$ & Input sequence of the recurrence, $\vy=\mW^{(L)}_\gamma\ve$ & Prop.~\ref{prop:recurrence} \\
$\vy$, $y_j$ & Output sequence of the operator & Prop.~\ref{prop:recurrence} \\
$\sigma_j$ & Innovation scale, $[\mW^{(L)}_\gamma]_{jj}$ & Prop.~\ref{prop:recurrence} \\
$\vs_j$ & Pre-input SSM state, containing only inputs before frame $j$ & \eqref{eq:app-ssm} \\
$\hat{\vs}_j$ & State estimate given $y_0,\dots,y_{j-1}$ & \eqref{eq:dssd-recurrence} \\
$\mSigma$ & Stationary state covariance & App.~\ref{app:recurrence} \\
$\mS_j$ & Conditional state covariance given $y_0,\dots,y_{j-1}$ & \eqref{eq:app-kalman} \\
$\vk_j$ & Kalman gain & \eqref{eq:app-kalman} \\
$D$ & Direct current-input term of the realization, $\sqrt\gamma$ & \eqref{eq:app-ssm} \\
\midrule
\multicolumn{3}{@{}l}{\emph{Training and sampling}}\\
$\mathcal{L}(\theta,\phi)$ & Mixed-noise regression loss & \eqref{eq:dssd-loss} \\
$\hat{\bm{\epsilon}}_c$ & Denoiser prediction of the mixed noise, $f_\theta(\vz_t,\vo,t)$ & \eqref{eq:dssd-ddim} \\
$\hat{\vx}$, $\hat{\bm{\epsilon}}$ & Estimates of the clean future and of the white noise at step $t$ & \eqref{eq:dssd-ddim} \\
$\bm{\xi}$ & White Gaussian noise that initializes sampling, $\vz_T=\mW^{(M)}_{\bar\gamma_T}\bm{\xi}$ & \S\ref{sec:method-training} \\
$\vv_t$ & Velocity field transported by the sampler & Prop.~\ref{prop:sampler} \\
\midrule
\multicolumn{3}{@{}l}{\emph{Analysis (Proposition~\ref{prop:information})}}\\
$\vz_t^{\mathrm{std}}$ & Standard corruption \eqref{eq:forward} at the same noise level & \S\ref{sec:method-analysis} \\
$\mF_t$ & $(\mW^{(M)}_{\bar\gamma_t})^{-1}\mW^{ff}_{\bar\gamma_t}$, future signal in whitened noise coordinates & \S\ref{sec:method-analysis} \\
$g$ & Estimator of $\vx$ & \eqref{eq:information} \\
\midrule
\multicolumn{3}{@{}l}{\emph{Appendix-only notation}}\\
$\nu$ & $\sqrt{1-\gamma}$ & App.~\ref{app:proofs} \\
$R_n$, $\vartheta_n$ & Pole modulus $\nu^{a_n}$ and angle $\Delta_\gamma\omega_n$ of mode $n$ & App.~\ref{app:proofs} \\
$Z$ & Normalizer of the kernel, $(\sum_\ell\tilde u_\ell^2)^{1/2}\ge1$ & App.~\ref{app:closed-form} \\
$V_\delta$ & Unnormalized kernel autocorrelation, $\sum_\ell\tilde u_\ell\tilde u_{\ell+\delta}$ & App.~\ref{app:closed-form} \\
$\mathcal{C}(R,\vartheta,\psi)$ & Closed form of $\sum_\ell R^\ell\cos(\ell\vartheta+\psi)$ & \eqref{eq:app-series} \\
$\varepsilon_j$ & Scalar i.i.d.\ $\mathcal{N}(0,1)$ inputs of the stationary process, $j\in\mathbb{Z}$ & App.~\ref{app:admissible} \\
$\mE$ & Matrix of the EMA recurrence & Cor.~\ref{cor:ema} \\
$\tilde{\mW}$ & Lower-triangular matrix of the unrolled recurrence & App.~\ref{app:recurrence} \\
$\beta_t$ & Noise variance $1-\bar\alpha_t$ & App.~\ref{app:sampler} \\
$\mB_t$ & Noise operator $\mW^{(M)}_{\bar\gamma_t}$ & App.~\ref{app:sampler} \\
$\vm_t$, $\mM_t$, $\mC_t$ & Coefficients of $\vz_t=\vm_t+\mM_t\vx+\mC_t\bm{\epsilon}$ & \eqref{eq:app-interpolant} \\
$f^\star$ & Unrestricted minimizer of \eqref{eq:dssd-loss} & App.~\ref{app:sampler} \\
$\chi$ & Smooth, compactly supported test function & App.~\ref{app:sampler} \\
$\mU$ & $(\mW^{(M)}_{\bar\gamma_t})^{-1}\mW^{fo}_{\bar\gamma_t}$ & App.~\ref{app:information} \\
$\bm{\zeta}$ & Auxiliary white noise independent of all other variables & App.~\ref{app:information} \\
$\bm{\mu}(\vx)$ & Mean of $q(\vz_t\mid\vo,\vx)$ & App.~\ref{app:score} \\
\midrule
\multicolumn{3}{@{}l}{\emph{Shorthand in Algorithms~\ref{alg:dssd-training} and~\ref{alg:dssd-sampling}}}\\
$\mH$, $\mG$, $\mB$ & $\mW^{fo}_\gamma$, $\mW^{ff}_\gamma$ and $\mW^{(M)}_\gamma$ returned by $\operatorname{BuildOperators}(\gamma)$ & App.~\ref{app:algorithms} \\
$\mathcal{E}$, $\mathcal{D}$ & Frozen encoder and decoder of the autoencoder & App.~\ref{app:algorithms} \\
$N_{\mathrm{b}}$ & Minibatch size & Alg.~\ref{alg:dssd-training} \\
$\tilde{\vx}_i$, $\tilde{\bm\epsilon}_i$ & Mixed signal and mixed noise of training example $i$ & Alg.~\ref{alg:dssd-training} \\
$t_S>\dots>t_0$ & Sampling grid of $S$ denoising steps, $t_0=0$ & Alg.~\ref{alg:dssd-sampling} \\
\midrule
\multicolumn{3}{@{}l}{\emph{Indices}}\\
$j$ & Frame (row) index within a window & \\
$k$ & Delay of an input in the impulse response & \\
$\ell$ & Delay of an input in the memory kernel & \\
$\delta$ & Lag between two frames & \\
$n$, $m$ & Mode indices & \\
\end{longtable}
}

\subsection{Closed-form sums}
\label{app:closed-form}

For $0\le R<1$ and real $\vartheta,\psi$, the real part of the geometric series
$\sum_{\ell\ge0}R^\ell e^{i(\ell\vartheta+\psi)}=e^{i\psi}/(1-R e^{i\vartheta})$ gives
\begin{equation}
\mathcal{C}(R,\vartheta,\psi)=\sum_{\ell\ge0}R^\ell\cos(\ell\vartheta+\psi)
=\frac{\cos\psi-R\cos(\psi-\vartheta)}{(1-R)^2+4R\sin^2(\vartheta/2)},
\label{eq:app-series}
\end{equation}
where the denominator equals $1-2R\cos\vartheta+R^2$ and is written in a form that avoids
cancellation. Using $\cos x\cos y=\frac12[\cos(x-y)+\cos(x+y)]$ with $x=\ell\vartheta_n$ and
$y=(\ell+\delta)\vartheta_m$,
\begin{align}
Z^2&=\sum_{\ell\ge0}\tilde u_\ell^2
=\frac12\sum_{n,m}r_nr_m\Big[\mathcal{C}(R_nR_m,\vartheta_n-\vartheta_m,0)
+\mathcal{C}(R_nR_m,\vartheta_n+\vartheta_m,0)\Big],\\
V_\delta&=\sum_{\ell\ge0}\tilde u_\ell\,\tilde u_{\ell+\delta}
=\frac12\sum_{n,m}r_nr_mR_m^{\delta}\Big[\mathcal{C}(R_nR_m,\vartheta_n-\vartheta_m,-\delta\vartheta_m)
+\mathcal{C}(R_nR_m,\vartheta_n+\vartheta_m,\delta\vartheta_m)\Big].
\end{align}
For $\delta\ge1$, the definition of $\kappa_\delta$ in \eqref{eq:dssd-operator} splits into the term $w_0w_\delta$ and the
terms with $k\ge1$, and $\sum_{k\ge1}w_kw_{k+\delta}=(1-\gamma)\sum_{\ell\ge0}u_\ell u_{\ell+\delta}$.
Since $u_\ell=\tilde u_\ell/Z$, this gives
\begin{equation}
\kappa_0=1,\qquad
\kappa_\delta=\sqrt{\gamma(1-\gamma)}\,u_{\delta-1}+(1-\gamma)\,V_\delta/Z^2
\quad(\delta\ge1).
\label{eq:app-kappa}
\end{equation}
Because the amplitudes are positive and sum to one, $\tilde u_0=1$ and $Z\ge1$.
As $\gamma\to1$, every $R_n\to0$, so $\tilde u_\ell\to0$ for $\ell>0$ while
$\tilde u_0=1$, giving the endpoint $w=(1,0,0,\dots)$.

\subsection{Admissibility (Proposition~\ref{prop:admissible})}
\label{app:admissible}

\begin{proof}[Proof of Proposition~\ref{prop:admissible}]
Let $(\varepsilon_j)_{j\in\mathbb{Z}}$ be i.i.d.\ $\mathcal{N}(0,1)$ and
$y_j=\sum_{k\ge0}w_k\varepsilon_{j-k}$. Since $\sum_kw_k^2=1$, the series converges in mean
square, $(y_j)$ is a stationary Gaussian process, and $\Cov(y_i,y_j)=\kappa_{|i-j|}$; hence
$\mJ^{(L)}$ is the covariance of $(y_0,\dots,y_{L-1})$.

(a) For $0\le j<L$, split $y_j=\sum_{k=0}^{j}w_{j-k}\varepsilon_k+y^{\mathrm{pre}}_j$, where
$y^{\mathrm{pre}}_j$ depends only on $(\varepsilon_k)_{k<0}$. The first term equals
$\mQ\,(\varepsilon_0,\dots,\varepsilon_{L-1})^\top$ with $\mQ$ lower triangular and diagonal
entries $w_0=\sqrt\gamma>0$, and it is independent of $y^{\mathrm{pre}}$. Thus
$\mJ^{(L)}=\mQ\mQ^\top+\Cov(y^{\mathrm{pre}})\succeq\mQ\mQ^\top\succ0$. A positive definite matrix
has a unique Cholesky factor, which is lower triangular with a positive diagonal and therefore
causal and invertible.

(b) $\operatorname{diag}(\mW\mW^\top)=\operatorname{diag}(\mJ)=\kappa_0=\sum_kw_k^2=1$.

(c) A conditional variance of a Gaussian vector depends only on its covariance, so we may
compute it for $(y_0,\dots,y_{L-1})^\top=\mW\ve$ with $\ve\sim\mathcal{N}(\bm{0},\mI)$. Because
$\mW$ is lower triangular with a positive diagonal, $(e_0,\dots,e_{j-1})$ is an invertible
linear function of $(y_0,\dots,y_{j-1})$, and $e_j$ is independent of it. Hence
$\operatorname{Var}(y_j\mid y_0,\dots,y_{j-1})=[\mW]_{jj}^2$. For the process itself,
$y_0,\dots,y_{j-1}$ are functions of $(\varepsilon_k)_{k\le j-1}$, whereas
$y_j=w_0\varepsilon_j+y^{\mathrm{past}}_j$ with $y^{\mathrm{past}}_j$ a function of $(\varepsilon_k)_{k\le j-1}$ and
$\varepsilon_j$ independent of the past. For jointly Gaussian variables, conditioning on more
information cannot increase the conditional variance, so
$[\mW]_{jj}^2=\operatorname{Var}(y_j\mid y_{<j})
\ge\operatorname{Var}\big(y_j\mid(\varepsilon_k)_{k\le j-1}\big)=w_0^2=\gamma$. Finally,
$\bar\gamma_t=1-\eta(1-\bar\alpha_t)\ge1-\eta$.
This gives a positive uniform lower bound if $\eta<1$. For $\eta=1$, each finite step
with $\bar\alpha_t>0$ remains admissible, but the zero-signal endpoint would have
$\bar\gamma_T=0$ and lies outside the construction.

(d) The leading $N\times N$ block of the Toeplitz matrix $\mJ^{(N+M)}$ is $\mJ^{(N)}$. In the
block form \eqref{eq:block}, $\mW^{oo}\mW^{oo\top}=\mJ^{(N)}$ and $\mW^{oo}$ is lower triangular
with a positive diagonal, so $\mW^{oo}=\mW^{(N)}$ by uniqueness. The trailing $M\times M$
block of $\mJ^{(N+M)}$ is likewise $\mJ^{(M)}$; multiplying out the block form yields
\begin{equation}
\mW^{fo}\mW^{fo\top}+\mW^{ff}\mW^{ff\top}=\mJ^{(M)}=\mW^{(M)}\mW^{(M)\top}.
\label{eq:app-schur}
\end{equation}
Moreover, the first row of $\mW^{(M)}$ is $(1,0,\dots,0)$ because $\kappa_0=1$, so the noise
branch of \eqref{eq:dssd-forward} leaves its first frame unchanged.

(e) At $\gamma=1$, $w=(1,0,\dots)$ by Appendix~\ref{app:closed-form}, so
$\mJ^{(L)}=\mI$ and $\mW^{(L)}=\mI$. As $\gamma\to1$, both coefficients in \eqref{eq:app-kappa}
tend to zero, while $|u_{\delta-1}|\le1$ and $|V_\delta/Z^2|=|\sum_\ell u_\ell u_{\ell+\delta}|\le1$
by the unit norm of $u$ and the Cauchy--Schwarz inequality. Hence $\kappa_\delta\to0$ for every
$\delta\ge1$, so $\mJ^{(L)}_\gamma\to\mI$, and continuity of the Cholesky factor on positive definite
matrices gives $\mW^{(L)}_\gamma\to\mI$.
\end{proof}

\begin{corollary}[EMA special case]
\label{cor:ema}
If $P=1$, $a_1=1$ and $\omega_1=0$, then for every $\gamma\in(0,1]$ the operator
$\mW^{(L)}_\gamma$ is the matrix of the recurrence $y_0=e_0$, $y_j=\nu\,y_{j-1}+\sqrt\gamma\,e_j$,
i.e., the exponential moving average operator of DyDiff \citep{guo2025dydiff}.
\end{corollary}

\begin{proof}
At $\gamma=1$ both matrices are the identity. For $\gamma<1$, $R_1=\nu$, $\tilde u_\ell=\nu^\ell$,
$Z^2=1/(1-\nu^2)=1/\gamma$, $u_\ell=\sqrt\gamma\,\nu^\ell$ and $w_k=\sqrt\gamma\,\nu^k$, so
$\kappa_\delta=\gamma\sum_k\nu^{2k+\delta}=\nu^\delta$. Let $\mE$ be the matrix of the recurrence,
with $E_{j0}=\nu^j$ and $E_{jk}=\sqrt\gamma\,\nu^{j-k}$ for $1\le k\le j$. For $i\ge j$,
\begin{equation*}
(\mE\mE^\top)_{ij}=\nu^{i+j}+\gamma\sum_{k=1}^{j}\nu^{i+j-2k}
=\nu^{i+j}+\nu^{i-j}\big(1-\nu^{2j}\big)=\nu^{i-j}=\kappa_{i-j},
\end{equation*}
using $\gamma=1-\nu^2$. Since $\mE$ is lower triangular with a positive diagonal,
$\mE=\operatorname{chol}(\mJ^{(L)})$.
\end{proof}

\subsection{State-space recurrence (Proposition~\ref{prop:recurrence})}
\label{app:recurrence}

\paragraph{State-space realization.}
Unlike the post-input state $\vh_k$ in \eqref{eq:ssm-recurrence}, we use a pre-input state
$\vs_j$ that contains only inputs before frame $j$. This convention implements the one-frame
delay in \eqref{eq:dssd-impulse}, with a separate direct term for the current input.
For $\gamma\in(0,1)$, define the $2P$-dimensional real system
\begin{equation}
\begin{aligned}
\bar{\mA}&=\operatorname{blockdiag}_{n}\bigg(R_n\begin{bmatrix}\cos\vartheta_n&-\sin\vartheta_n\\
\sin\vartheta_n&\cos\vartheta_n\end{bmatrix}\bigg),\qquad D=\sqrt\gamma,\\
\vb&=(1,0,\dots,1,0)^\top,\qquad
\vc=\frac{\sqrt{1-\gamma}}{Z}\,(r_1,0,\dots,r_P,0)^\top .
\end{aligned}
\end{equation}
Each block rotates $(1,0)^\top$ by $\ell\vartheta_n$ and scales it by $R_n^\ell$, so
$\vc^\top\bar{\mA}^{\ell}\vb=\sqrt{1-\gamma}\,u_\ell=w_{\ell+1}$. Hence, with
$\vs_j=\sum_{k\ge1}\bar{\mA}^{k-1}\vb\,\varepsilon_{j-k}$, the process of
Appendix~\ref{app:admissible} satisfies
\begin{equation}
\vs_{j+1}=\bar{\mA}\vs_j+\vb\,\varepsilon_j,
\qquad
y_j=\vc^\top\vs_j+D\,\varepsilon_j .
\label{eq:app-ssm}
\end{equation}
Because the spectral radius $\max_nR_n$ is below one, $\vs_j$ is stationary with covariance
$\mSigma$ solving $\mSigma=\bar{\mA}\mSigma\bar{\mA}^\top+\vb\vb^\top$, and $\vs_j$ is
independent of $(\varepsilon_k)_{k\ge j}$.

\paragraph{Innovations.}
Let $\hat{\vs}_j=\E[\vs_j\mid y_0,\dots,y_{j-1}]$ and
$\mS_j=\Cov(\vs_j\mid y_0,\dots,y_{j-1})$, so $\hat{\vs}_0=\bm{0}$ and $\mS_0=\mSigma$.
Conditional on $y_{<j}$, the pair $(\vs_{j+1},y_j)$ is Gaussian with means
$(\bar{\mA}\hat{\vs}_j,\vc^\top\hat{\vs}_j)$, variance
$\sigma_j^2=\operatorname{Var}(y_j\mid y_{<j})=\vc^\top\mS_j\vc+D^2$ and cross-covariance
$\Cov(\vs_{j+1},y_j\mid y_{<j})=\bar{\mA}\mS_j\vc+\vb D$, because $\varepsilon_j$ is
independent of $\vs_j$ and of $y_{<j}$. Gaussian conditioning gives the Kalman recursion
\begin{equation}
\vk_j=\frac{\bar{\mA}\mS_j\vc+\vb D}{\sigma_j^2},\qquad
\hat{\vs}_{j+1}=\bar{\mA}\hat{\vs}_j+\vk_j\,\big(y_j-\vc^\top\hat{\vs}_j\big),\qquad
\mS_{j+1}=\bar{\mA}\mS_j\bar{\mA}^\top+\vb\vb^\top-\sigma_j^2\,\vk_j\vk_j^\top .
\label{eq:app-kalman}
\end{equation}
Let $e_j=(y_j-\vc^\top\hat{\vs}_j)/\sigma_j$. By the defining property of conditional
expectation, $e_j$ is uncorrelated with every function of $y_{<j}$, and each $e_k$ with $k<j$
is such a function. The $e_j$ are therefore uncorrelated, jointly Gaussian and of unit
variance, hence i.i.d.\ $\mathcal{N}(0,1)$. Substituting $y_j=\vc^\top\hat{\vs}_j+\sigma_j\,e_j$
into \eqref{eq:app-kalman} gives the recurrence of Proposition~\ref{prop:recurrence},
\begin{equation}
y_j=\vc^\top\hat{\vs}_j+\sigma_j\,e_j,
\qquad
\hat{\vs}_{j+1}=\bar{\mA}\hat{\vs}_j+\sigma_j\,\vk_j\,e_j,
\qquad \hat{\vs}_0=\bm{0},
\label{eq:dssd-recurrence}
\end{equation}
whose input sequence is $\ve=(e_0,\dots,e_{L-1})$.
Both $\vs_j$ and its estimate $\hat{\vs}_j$ precede the current output; the latter is initialized
at zero mean with uncertainty $\mS_0=\mSigma$, not at a known zero physical state.

\begin{proof}[Proof of Proposition~\ref{prop:recurrence}]
Unrolling \eqref{eq:dssd-recurrence} gives $(y_0,\dots,y_{L-1})^\top=\tilde{\mW}\,(e_0,\dots,e_{L-1})^\top$
for a lower-triangular $\tilde{\mW}$ with diagonal entries $\sigma_j\ge D=\sqrt\gamma>0$. Then
$\mJ^{(L)}=\Cov(y)=\tilde{\mW}\tilde{\mW}^\top$, and uniqueness of the Cholesky factor gives
$\tilde{\mW}=\mW^{(L)}$, in particular $\sigma_j=[\mW^{(L)}]_{jj}$. Since the recurrence is linear and
its coefficients do not depend on the input, it computes $\mW^{(L)}\ve$ for any input sequence $\ve$.
Furthermore, $\sigma_0^2=\vc^\top\mSigma\vc+D^2=\operatorname{Var}(y_0)=1$, and by stationarity
$\sigma_j^2=\operatorname{Var}(y_0\mid y_{-j},\dots,y_{-1})$, which cannot increase as the
conditioning set grows.
\end{proof}

In the setting of Corollary~\ref{cor:ema}, $y_0$ determines the state, so $\mS_1=\bm{0}$ and the
recurrence closes on the previous output. The gains depend only on $(\gamma,\phi,j)$ and cost
$O(LP^3)$ per dynamics level, after which applying the recurrence to $d$-dimensional frames
costs $O(LPd)$, compared with $O(L^3+L^2d)$ for \eqref{eq:dssd-operator}.

\subsection{Strictly richer correlations (Proposition~\ref{prop:richer})}
\label{app:richer}

\begin{proof}[Proof of Proposition~\ref{prop:richer}]
The EMA special case follows from Corollary~\ref{cor:ema}, whose proof gives
$\kappa_\delta=\nu^\delta=(1-\gamma)^{\delta/2}>0$. To exhibit a correlation outside that family,
fix $\gamma\in(0,1)$, take $P=1$ and $\omega_1=\pi/\Delta$, so that $\vartheta_1=\pi$, and write
$R=R_1=\nu^{a_1}\in(0,1)$. Then
$\tilde u_\ell=R^\ell\cos(\ell\pi)=(-R)^\ell$, $Z^2=\sum_\ell R^{2\ell}=1/(1-R^2)$ and
$u_\ell=\sqrt{1-R^2}\,(-R)^\ell$, so
\begin{equation*}
\sum_{\ell\ge0}u_\ell u_{\ell+1}=(1-R^2)\sum_{\ell\ge0}(-R)^{2\ell+1}
=-R\,(1-R^2)\sum_{\ell\ge0}R^{2\ell}=-R .
\end{equation*}
By \eqref{eq:app-kappa} with $V_1/Z^2=\sum_\ell u_\ell u_{\ell+1}$ and $u_0=\sqrt{1-R^2}$,
\begin{equation*}
\kappa_1=\sqrt{\gamma(1-\gamma)}\,\sqrt{1-R^2}-(1-\gamma)\,R
=\sqrt{1-\gamma}\,\Big(\sqrt{\gamma}\sqrt{1-R^2}-\sqrt{1-\gamma}\,R\Big),
\end{equation*}
which is negative if and only if $(1-\gamma)R^2>\gamma(1-R^2)$, that is, $R^2>\gamma$. Since
$R^2=(1-\gamma)^{a_1}$ and $\log(1-\gamma)<0$, this holds exactly when
$a_1<\log\gamma/\log(1-\gamma)$, and such $a_1>0$ exists because the right-hand side is positive.
\end{proof}

For example, $\gamma=0.5$ and $a_1=0.5$ give $\kappa_1\approx-0.150$. The required frequency
$\omega_1=\pi/\Delta_\gamma$ lies within the clipping range $[0,2\pi]$ of
Appendix~\ref{app:practical} whenever $\Delta_\gamma\ge\frac12$, i.e.,
$\gamma\ge1-e^{-1/2}\approx0.39$, which covers every $\bar\gamma_t$ when $\eta\le e^{-1/2}$,
including $\eta=0.5$.

\subsection{Sampler consistency (Proposition~\ref{prop:sampler})}
\label{app:sampler}

Write $\beta_t=1-\bar\alpha_t$, $\mB_t=\mW^{(M)}_{\bar\gamma_t}$,
$\vm_t=\sqrt{\bar\alpha_t}\,\mW^{fo}_{\bar\gamma_t}\vo$, $\mM_t=\sqrt{\bar\alpha_t}\,\mW^{ff}_{\bar\gamma_t}$
and $\mC_t=\sqrt{\beta_t}\,\mB_t$, so that \eqref{eq:dssd-forward} reads
\begin{equation}
\vz_t=\vm_t+\mM_t\vx+\mC_t\bm{\epsilon},
\label{eq:app-interpolant}
\end{equation}
with the same $(\vx,\bm{\epsilon})$ for every $t$. For $0<\bar\alpha_t<1$, $\mM_t$ and $\mC_t$ are
invertible by Proposition~\ref{prop:admissible}(a).

\begin{proof}[Proof of Proposition~\ref{prop:sampler}]
\emph{Posterior means.} The minimizer of \eqref{eq:dssd-loss} at step $t$ is
$f^\star=\E[\mB_t\bm{\epsilon}\mid\vz_t,\vo]$. Taking the conditional expectation of
\eqref{eq:app-interpolant} given $(\vz_t,\vo)$, and using $\E[\vz_t\mid\vz_t,\vo]=\vz_t$, yields
$\vz_t=\vm_t+\mM_t\E[\vx\mid\vz_t,\vo]+\sqrt{\beta_t}\,f^\star$. Solving for the first
expectation gives exactly $\hat{\vx}$ in \eqref{eq:dssd-ddim} with $\hat{\bm{\epsilon}}_c=f^\star$,
and $\E[\bm{\epsilon}\mid\vz_t,\vo]=\mB_t^{-1}f^\star=\hat{\bm{\epsilon}}$. By linearity of
\eqref{eq:app-interpolant} at step $t'$,
$\E[\vz_{t'}\mid\vz_t,\vo]=\vm_{t'}+\mM_{t'}\hat{\vx}+\mC_{t'}\hat{\bm{\epsilon}}$, which is the update
\eqref{eq:dssd-ddim}.

\emph{Transport.} The entries of $\mJ^{(L)}_\gamma$ are smooth in $\gamma\in(0,1)$ by
Appendix~\ref{app:closed-form}, the Cholesky factor is smooth on positive definite matrices, and
$\bar\gamma_t$ is affine in $\bar\alpha_t$; hence $\vm_t$, $\mM_t$ and $\mC_t$ are twice
continuously differentiable on interior time intervals for $\eta>0$. For $\eta=0$, the
operator is identically $\mI$ and the same conclusion follows directly. Define
$\vv_t(\vz)=\dot{\vm}_t+\dot{\mM}_t\E[\vx\mid\vz_t=\vz,\vo]+\dot{\mC}_t\E[\bm{\epsilon}\mid\vz_t=\vz,\vo]$.
For a smooth, compactly supported test function $\chi$ and data with finite second moments,
\begin{equation*}
\frac{d}{dt}\E\big[\chi(\vz_t)\mid\vo\big]
=\E\big[\nabla\chi(\vz_t)^\top(\dot{\vm}_t+\dot{\mM}_t\vx+\dot{\mC}_t\bm{\epsilon})\,\big|\,\vo\big]
=\E\big[\nabla\chi(\vz_t)^\top\vv_t(\vz_t)\,\big|\,\vo\big],
\end{equation*}
by the tower property, which is the weak form of the continuity equation
$\partial_tq_t+\nabla\cdot(q_t\vv_t)=0$ for $q_t=q(\vz_t\mid\vo)$.

\emph{Consistency.} By the first part, $\vz_t=\vm_t+\mM_t\hat{\vx}+\mC_t\hat{\bm{\epsilon}}$, so
$\vz_{t'}-\vz_t=(\vm_{t'}-\vm_t)+(\mM_{t'}-\mM_t)\hat{\vx}+(\mC_{t'}-\mC_t)\hat{\bm{\epsilon}}
=(t'-t)\,\vv_t(\vz_t)+O\big((t'-t)^2\big)$ by Taylor's theorem, where the remainder is uniform
wherever the posterior means are bounded.
\end{proof}

Assuming the velocity admits a unique flow and the continuity equation has a unique
solution in the considered class, integrating
$\dot{\vz}=\vv_t(\vz)$ backward from the exact terminal marginal reproduces every marginal
$q_t$ on the interior interval, and \eqref{eq:dssd-ddim} is its first-order discretization.
Extension to a degenerate endpoint requires separate control of the flow limit.
At the last step,
$\bar\alpha_{t'}=\bar\gamma_{t'}=1$ gives $\vm_{t'}=\bm{0}$, $\mM_{t'}=\mI$ and $\mC_{t'}=\bm{0}$, so the
update returns $\hat{\vx}=\E[\vx\mid\vz_t,\vo]$. Two approximations remain, and both are shared
with DDIM. First, by the law of total covariance, $\Cov(\vz_{t'}\mid\vo)$ exceeds the covariance of
$\E[\vz_{t'}\mid\vz_t,\vo]$ by $\E[\Cov(\vz_{t'}\mid\vz_t,\vo)\mid\vo]\succeq0$, so a large step omits
conditional variance. Second, the initialization $\mB_T\bm{\xi}$ matches $q_T$ for arbitrary
data distributions at $\bar\alpha_T=0$, provided $\bar\gamma_T>0$, since then
$\vm_T=\bm{0}$ and $\mM_T=\bm{0}$. At $\bar\alpha_T>0$ it is generally approximate;
special data distributions can still yield equality. These statements concern the ideal
denoiser under the forward marginals, not errors along a learned sampler's trajectories.

\subsection{Information ordering (Proposition~\ref{prop:information})}
\label{app:information}

\begin{proof}[Proof of Proposition~\ref{prop:information}]
Fix $t$, write $\mB=\mW^{(M)}_{\bar\gamma_t}$, $\mF=\mB^{-1}\mW^{ff}$ and $\mU=\mB^{-1}\mW^{fo}$, and
let $\beta=1-\bar\alpha_t$. Multiplying \eqref{eq:app-schur} by $\mB^{-1}$ on the left and by
$\mB^{-\top}$ on the right gives $\mF\mF^\top+\mU\mU^\top=\mI$, hence
$\mF\mF^\top=\mI-\mU\mU^\top\preceq\mI$.

Given $\vo$, the map $\vz_t\mapsto\tilde{\vz}_t=\mB^{-1}(\vz_t-\sqrt{\bar\alpha_t}\,\mW^{fo}\vo)
=\sqrt{\bar\alpha_t}\,\mF\vx+\sqrt\beta\,\bm{\epsilon}$ is invertible, so estimators based on
$(\vz_t,\vo)$ and on $(\tilde{\vz}_t,\vo)$ achieve the same errors. Let
$\bm{\zeta}\sim\mathcal{N}(\bm{0},\mI)$ be independent of everything else and define
$\tilde{\vz}'_t=\mF\vz_t^{\mathrm{std}}+\sqrt\beta\,(\mI-\mF\mF^\top)^{1/2}\bm{\zeta}$. Conditional on
$(\vx,\vo)$, $\tilde{\vz}'_t$ is Gaussian with mean $\sqrt{\bar\alpha_t}\,\mF\vx$ and covariance
$\beta\,\mF\mF^\top+\beta\,(\mI-\mF\mF^\top)=\beta\mI$, which is the conditional law of
$\tilde{\vz}_t$. Hence $(\vx,\vo,\tilde{\vz}'_t)$ and $(\vx,\vo,\tilde{\vz}_t)$ have the same joint
distribution, and $\tilde{\vz}'_t$ is computed from $\vz_t^{\mathrm{std}}$ and $\bm{\zeta}$ without
$\vx$. For any estimator $g$ with finite risk, $g(\tilde{\vz}'_t,\vo)$ is a function of
$(\vz_t^{\mathrm{std}},\bm{\zeta},\vo)$, so
$\E\lVert\vx-g(\tilde{\vz}_t,\vo)\rVert^2=\E\lVert\vx-g(\tilde{\vz}'_t,\vo)\rVert^2
\ge\E\lVert\vx-\E[\vx\mid\vz_t^{\mathrm{std}},\bm{\zeta},\vo]\rVert^2
=\E\lVert\vx-\E[\vx\mid\vz_t^{\mathrm{std}},\vo]\rVert^2$, where the inequality holds because the
conditional mean minimizes the squared error for data with finite second moments, and the last
equality uses the independence of $\bm{\zeta}$. Minimizing over $g$ yields \eqref{eq:information}.
\end{proof}

For the EMA operator of Corollary~\ref{cor:ema} with $N\ge1$, every future column of
$\mW^{(N+M)}$ has global index at least one, so its entries are $\sqrt\gamma\,\nu^{j-k}$, whereas
$\mW^{(M)}$ has first column $(\nu^{j})_j$ and the same entries elsewhere. Hence
$\mW^{ff}=\mW^{(M)}\operatorname{diag}(\sqrt\gamma,1,\dots,1)$ and
$\mF=\operatorname{diag}(\sqrt\gamma,1,\dots,1)$: only the first future frame is attenuated. For a
general DSSD operator, $\mI-\mF\mF^\top=\mU\mU^\top$ has rank at most $\min(N,M)$.

\subsection{Colored-noise regression and the score}
\label{app:score}

Fix $t$ and $\phi$, and write $\beta=1-\bar\alpha_t$, $\mB=\mW^{(M)}_{\bar\gamma_t}$ and
$\bm{\mu}(\vx)=\sqrt{\bar\alpha_t}\,(\mW^{fo}\vo+\mW^{ff}\vx)$. For each $t$, the minimizer of
\eqref{eq:dssd-loss} over unrestricted functions is
$f^\star(\vz,\vo,t)=\E[\mB\bm{\epsilon}\mid\vz_t=\vz,\vo]$. The marginal is the mixture
$q(\vz\mid\vo)=\int p(\vx\mid\vo)\,\mathcal{N}\big(\vz;\bm{\mu}(\vx),\beta\mB\mB^\top\big)\,d\vx$.
Differentiating under the integral, which is valid under standard integrability conditions,
\begin{equation}
\nabla_{\vz}\log q(\vz\mid\vo)
=\E\big[-(\beta\mB\mB^\top)^{-1}\big(\vz-\bm{\mu}(\vx)\big)\,\big|\,\vz_t=\vz,\vo\big]
=-\beta^{-1/2}\,(\mB\mB^\top)^{-1}f^\star(\vz,\vo,t),
\end{equation}
since $\vz_t-\bm{\mu}(\vx)=\sqrt\beta\,\mB\bm{\epsilon}$. For any fixed operator, the colored
target therefore determines the score, and the correlation between target coordinates requires
no whitening of \eqref{eq:dssd-loss}. Whitening would regress on $\mB^{-1}$ times the target,
that is, on $\bm{\epsilon}$ itself; since conditional expectation commutes with the fixed matrix
$\mB$, the unrestricted minimizers of the two losses are related by
$\E[\bm{\epsilon}\mid\vz_t,\vo]=\mB^{-1}f^\star$, so they carry the same information, and the
unwhitened minimizer is the one that enters the sampling update \eqref{eq:dssd-ddim} directly.
The two losses do weight the coordinates of the residual differently, which can matter for a
finite denoiser; this is part of the regression geometry discussed in
Section~\ref{sec:method-analysis}. This statement concerns a fixed $\phi$; the value of $\phi$
that minimizes \eqref{eq:dssd-loss} need not minimize a forecasting error.

\subsection{Nonzero-frequency initialization}
\label{app:zero-frequency}

For $\ell\ge0$,
$\partial\tilde u_\ell/\partial\omega_n=-r_nR_n^\ell\,\ell\Delta\,\sin(\ell\Delta\omega_n)$, which
vanishes for every $\ell$ and $\gamma$ when $\omega_n=0$. The normalized kernel, the
autocorrelation, the operator \eqref{eq:dssd-operator} and the loss \eqref{eq:dssd-loss} depend
on $\omega_n$ only through $(\tilde u_\ell)_{\ell\ge0}$, so $\partial\mathcal{L}/\partial\omega_n=0$
at $\omega_n=0$. With a zero gradient, first-order optimizers such as Adam leave both the
parameter and its moment estimates unchanged, so a mode initialized at zero frequency remains
non-oscillatory throughout training. This motivates the nonzero initial frequencies of
Appendix~\ref{app:practical}.

\subsection{Stable parameterization and optimization}
\label{app:practical}

\paragraph{Bounded modes.}
Decay rates and frequencies are clipped to $\log a_n\in[-9,3]$ and $\omega_n\in[0,2\pi]$, and
amplitudes are kept positive by the softmax. The closed-form sums and the Cholesky
factorization are computed in double precision.

\paragraph{Initialization.}
Decay rates are log-spaced in $[0.12,3]$, frequencies are set to
$\omega_n=2\pi(n-\frac12)/P$, and amplitudes are uniform. Nonzero initial frequencies are
necessary by the argument of Appendix~\ref{app:zero-frequency}.

\paragraph{Operator certificates.}
Proposition~\ref{prop:admissible} holds exactly, but finite-precision inverses can still
amplify errors when a learned operator is ill conditioned. Before training and at every
evaluation, we therefore check the operator on a fixed grid of $33$ dynamics levels taken from
the schedule, at window lengths $M$ and $N+M$: the reconstruction error of the full and
history-conditioned inverses, unit variance, causality, the identity at $\gamma=1$, and a
condition number below $10^{3}$. The checks are run in single and double precision, and
training stops if any of them fails. These checks certify only the tested dynamics levels
and checkpoints, not every possible parameter value or every point between grid levels.
\clearpage
\raggedbottom
\section{Training and Sampling Algorithms}
\label{app:algorithms}

Algorithms~\ref{alg:dssd-training} and~\ref{alg:dssd-sampling} summarize the mixed-noise
training objective and deterministic sampler. The pretrained encoder $\mathcal{E}$ and decoder $\mathcal{D}$
are frozen. All temporal matrices act independently on each latent coordinate.

\paragraph{Shared operator construction.}
Both algorithms use $\operatorname{BuildOperators}(\gamma)$, with the current dynamics
parameters $\phi$ and window lengths $N,M$ understood. It returns three matrices:
$\mH$, which maps the observed history into the future; $\mG$, which mixes future signal
frames; and $\mB$, which mixes future noise. The routine constructs the normalized SSM
kernel and impulse response in \eqref{eq:dssd-kernel} and \eqref{eq:dssd-impulse}, computes
their stationary covariance matrices for lengths $N+M$ and $M$ using
Appendix~\ref{app:closed-form}, and takes their lower-triangular Cholesky factors as in
\eqref{eq:dssd-operator}. The blocks $\mH=\mW^{fo}_\gamma$ and $\mG=\mW^{ff}_\gamma$
come from the length-$N+M$ factor, while $\mB=\mW^{(M)}_\gamma$.
In general, $\mG\ne\mB$: signal mixing includes the observed history, whereas noise mixing
uses a separate future window. At $\gamma=1$, the routine returns $(\bm{0},\mI_M,\mI_M)$.
Parameter bounds and double-precision covariance calculations follow
Appendix~\ref{app:practical}. The construction is differentiable in $\phi$ during training.

\noindent
\begin{minipage}[t]{0.485\textwidth}
\vspace{0pt}
\begin{algorithm}[H]
\caption{Training of DSSD}
\label{alg:dssd-training}
\small
\algrenewcommand\algorithmicindent{0.8em}
\begin{algorithmic}[1]
\Require Training sequences; frozen encoder $\mathcal{E}$; window lengths $N,M$; batch size $N_{\mathrm{b}}$;
         schedule $\{\bar\alpha_t\}_{t=1}^{T}$; mixing strength $\eta$; optimizer and update budget
\Ensure Trained parameters $(\theta,\phi)$
\State Initialize $(\theta,\phi)$ as in Appendix~\ref{app:practical}
\While{update budget remains}
    \State Sample a minibatch $\{(O_i,X_i)\}_{i=1}^{N_{\mathrm{b}}}$
    \State $\vo_i\gets\mathcal{E}(O_i)$; $\vx_i\gets\mathcal{E}(X_i)$
    \For{$i=1,\ldots,N_{\mathrm{b}}$}
        \State Sample $t_i\sim\operatorname{Uniform}\{1,\ldots,T\}$
        \State Sample $\bm\epsilon_i\sim\mathcal N(\bm0,\mI)$
        \State $\gamma_i\gets1-\eta(1-\bar\alpha_{t_i})$
        \State $(\mH_i,\mG_i,\mB_i)\gets\operatorname{BuildOperators}(\gamma_i)$
        \State $\tilde{\vx}_i\gets\mH_i\vo_i+\mG_i\vx_i$
        \State $\tilde{\bm\epsilon}_i\gets\mB_i\bm\epsilon_i$
        \State $\vz_i\gets\sqrt{\bar\alpha_{t_i}}\,\tilde{\vx}_i+
               \sqrt{1-\bar\alpha_{t_i}}\,\tilde{\bm\epsilon}_i$
        \State $\hat{\bm\epsilon}_{c,i}\gets f_\theta(\vz_i,\vo_i,t_i)$
    \EndFor
    \State $\mathcal L\gets N_{\mathrm{b}}^{-1}\sum_{i=1}^{N_{\mathrm{b}}}\lVert\hat{\bm\epsilon}_{c,i}-\tilde{\bm\epsilon}_i\rVert_2^2$
    \State Compute $\nabla_{\theta,\phi}\mathcal L$ through $\vz_i$ and $\tilde{\bm\epsilon}_i$
    \State Update $(\theta,\phi)$
\EndWhile
\State \Return $(\theta,\phi)$
\end{algorithmic}
\end{algorithm}
\end{minipage}\hfill
\begin{minipage}[t]{0.485\textwidth}
\vspace{0pt}
\begin{algorithm}[H]
\caption{Sampling of DSSD}
\label{alg:dssd-sampling}
\small
\algrenewcommand\algorithmicindent{0.8em}
\begin{algorithmic}[1]
\Require Observed frames $O$; frozen encoder $\mathcal{E}$ and decoder $\mathcal{D}$; trained $(\theta,\phi)$;
         prediction length $M$; schedule $\bar\alpha$ and mixing strength $\eta$;
         grid $T\ge t_S>\cdots>t_1>t_0=0$
\Ensure Predicted future frames $\hat X$
\State $\vo\gets\mathcal{E}(O)$; $N\gets\text{number of frames in }\vo$
\State Set $\bar\alpha_0\gets1$ and $\bar\gamma_0\gets1$
\For{$j=0,\ldots,S$}
    \State $\bar\gamma_{t_j}\gets1-\eta(1-\bar\alpha_{t_j})$
    \State $(\mH_{t_j},\mG_{t_j},\mB_{t_j})\gets
           \operatorname{BuildOperators}(\bar\gamma_{t_j})$
\EndFor
\State Draw $\bm\xi\sim\mathcal N(\bm0,\mI)$ of shape $M\times d$
\State $\vz_{t_S}\gets\mB_{t_S}\bm\xi$
\For{$j=S,\ldots,1$}
    \State $t\gets t_j$; $t'\gets t_{j-1}$
    \State $\hat{\bm\epsilon}_c\gets f_\theta(\vz_t,\vo,t)$
    \State Solve $\mG_t\hat{\vx}=(\vz_t-\sqrt{1-\bar\alpha_t}\,\hat{\bm\epsilon}_c)/\sqrt{\bar\alpha_t}-\mH_t\vo$
           for $\hat{\vx}$
    \State Solve $\mB_t\hat{\bm\epsilon}=\hat{\bm\epsilon}_c$ for
           $\hat{\bm\epsilon}$
    \State $\vz_{t'}\gets\sqrt{\bar\alpha_{t'}}(\mH_{t'}\vo+\mG_{t'}\hat{\vx})$
    \State $\vz_{t'}\gets\vz_{t'}+\sqrt{1-\bar\alpha_{t'}}\,\mB_{t'}\hat{\bm\epsilon}$
\EndFor
\State \Return $\hat X\gets\mathcal{D}(\vz_0)$
\end{algorithmic}
\end{algorithm}
\end{minipage}

\paragraph{Training details.}
The noise target is not detached from the dynamics parameters. The loss is written as a
batch average of squared norms; the implementation additionally averages over latent
coordinates. Operator checks are performed before training and at evaluation checkpoints
as described in Appendix~\ref{app:practical}.

\paragraph{Sampling details.}
All model-evaluation steps satisfy $0<\bar\alpha_{t_j}<1$ and
$\bar\gamma_{t_j}>0$. The parameters remain fixed, so the operators can be precomputed;
no gradients or new noise draws are needed inside the sampling loop. At $t'=0$,
$(\mH_0,\mG_0,\mB_0)=(\bm0,\mI_M,\mI_M)$, and the update returns
$\vz_0=\hat{\vx}$. The initial colored Gaussian is generally an approximation to the
terminal forward marginal at nonzero $\bar\alpha_{t_S}$, as discussed in
Appendix~\ref{app:sampler}. We use steps $1,\ldots,T$ and the clean endpoint $0$;
the implementation uses array indices $0,\ldots,T-1$ and a clean-endpoint sentinel $-1$.

\section{Experimental Details}
\label{app:experiments}

\subsection{Data and forecasting tasks}
\label{app:experiments-data}

Table~\ref{tab:exp-data} specifies the input and prediction windows. All models forecast future frames from observed frames. We map SEVIR and RGB inputs from $[0,255]$ to $[-1,1]$ and divide both Turbulence velocity components by four, without clipping.

\begin{table}[ht]
\centering
\small
\setlength{\tabcolsep}{5pt}
\caption{Forecasting tasks. $N$ and $M$ are the numbers of observed and future frames.}
\label{tab:exp-data}
\begin{tabular}{l l c c l}
\hline
Dataset & Input & Resolution & $N\!\to\!M$ & Input transform \\
\hline
SEVIR & VIL, 1 channel & $128^2$ & $7\!\to\!6$ & $x\mapsto x/127.5-1$ \\
Turbulence & Velocity, 2 channels & $64^2$ & $4\!\to\!11$ & $x\mapsto x/4$ \\
BAIR & RGB video & $64^2$ & $1\!\to\!15$ & $x\mapsto x/127.5-1$ \\
RoboNet & RGB video & $64^2$ & $2\!\to\!10$ & $x\mapsto x/127.5-1$ \\
\hline
\end{tabular}
\end{table}

\subsection{Architecture, optimization, and sampling}
\label{app:experiments-training}

Within each dataset, all methods use the same pretrained, frozen, framewise VAE \citep{kingma2014vae}, including
its EMA weights. The VAE reduces each spatial dimension by a factor of four, outputs three
latent channels, and uses a latent scale factor of $0.18215$. During training, latent
frames are sampled from the VAE posterior; evaluation encodes observations using its
posterior mode. The U-Net takes clean observed latents and noisy future latents as
conditioning and prediction inputs, respectively. We use a video U-Net \citep{ronneberger2015unet} inspired by
Stable Video Diffusion \citep{blattmann2023svd}, configured identically for all
three methods.

\begin{table}[ht]
\centering
\small
\setlength{\tabcolsep}{6pt}
\renewcommand{\arraystretch}{1.3}
\caption{Shared model, optimization, and sampling settings.}
\label{tab:exp-shared}
\begin{tabular}{p{0.27\linewidth} p{0.62\linewidth}}
\hline
Component & Setting \\
\hline
Video U-Net & 64 base channels; channel multipliers $(1,2,4,4)$; two residual blocks per level. \\
Attention and time mixing & Attention at downsampling factors $(4,2,1)$; four heads; transformer depth 1; dropout 0; temporal convolution kernel $(3,1,1)$ and learned spatial/temporal mixing. \\
Diffusion & 1{,}000 timesteps; cosine schedule \citep{nichol2021improved} with $s=0.008$; $\epsilon$ prediction. \\
Optimization & Adam \citep{kingma2015adam}, $(\beta_1,\beta_2)=(0.9,0.999)$, $\epsilon_{\rm Adam}=10^{-8}$, weight decay 0; effective batch size 16; FP32. No denoiser-weight EMA. \\
Sampling & 50 deterministic DDIM steps on the same \texttt{released\_uniform} timestep grid. \\
\hline
\end{tabular}
\end{table}

\begin{table}[ht]
\centering
\small
\setlength{\tabcolsep}{5pt}
\renewcommand{\arraystretch}{1.3}
\caption{Method-specific temporal operator and learning-rate schedule.}
\label{tab:exp-methods}
\begin{tabular}{p{0.17\linewidth} p{0.42\linewidth} p{0.29\linewidth}}
\hline
Method & Temporal operator & Learning rate \\
\hline
DPM & Identity (no temporal mixing). & $10^{-4}$. \\
DyDiff & Fixed EMA, $\eta=0.5$. & $10^{-4}$. \\
DSSD & Learned stationary SSM operator, state size 12 (six conjugate pairs), $\eta=0.5$ & $10^{-4}$. \\
\hline
\end{tabular}
\end{table}

The DSSD operator parameters form a separate optimizer group; the learning-rate schedule applies to both groups. DSSD's stationary covariance and Cholesky factor are computed in double precision, with the parameter bounds and numerical operator checks detailed in Appendix~\ref{app:practical}.

\paragraph{DiT backbone.}
For the SEVIR backbone ablation (Appendix~\ref{app:additional-results-backbone}), we replace
the U-Net with a DiT-S/4 \citep{peebles2023dit}: 12 blocks of width 384 with six attention
heads, MLP ratio 4 and no dropout. Each latent frame is split spatially into $4\times4$
patches, giving 64 tokens per frame and 832 tokens for the 13 frames, and attention runs
jointly over all frames with fixed sine--cosine spatial and temporal position embeddings.
The inputs are the same as for the U-Net, the timestep enters through adaLN-Zero, and only
the outputs at future-frame positions are used. The DiT has 32{,}529{,}840 parameters; all
other settings, including the method-specific learning rates, match the U-Net runs on SEVIR.

\paragraph{Compute.}
Each model is trained on a single NVIDIA A100 80GB GPU, using the PCIe
variant for SEVIR and Turbulence and the SXM4 variant for BAIR and RoboNet.
Within each dataset, the three methods share the same GPU type and batch
configuration. DSSD adds 18 trainable parameters for its six conjugate mode
pairs to the 60{,}982{,}185-parameter denoiser shared by all methods, whereas
DyDiff's fixed EMA operator adds none. Table~\ref{tab:exp-compute} reports the
training cost of the first 50{,}000 updates. Relative to the faster baseline, DSSD's
median time per update is 1.4\% higher on SEVIR and 2.0\% higher on BAIR, is the lowest
on Turbulence, and is 4.9\% higher on RoboNet, computed from the values in
Table~\ref{tab:exp-compute}. All three methods reserve the same GPU memory on every
dataset. Learning the temporal operator therefore adds little training cost.

\begin{table}[ht]
\centering
\footnotesize
\setlength{\tabcolsep}{2pt}
\caption{Training cost over the first 50{,}000 updates on one A100 80GB GPU.
s/upd.: median time per update (s); Total: wall-clock time (h);
Mem.: reserved GPU memory (GiB).}
\label{tab:exp-compute}
\begin{tabular*}{\linewidth}{@{\extracolsep{\fill}}l*{12}{c}@{}}
\toprule
\multirow{2}{*}{Method} &
\multicolumn{3}{c}{SEVIR} & \multicolumn{3}{c}{Turbulence} &
\multicolumn{3}{c}{BAIR} & \multicolumn{3}{c}{RoboNet} \\
\cmidrule(lr){2-4}\cmidrule(lr){5-7}\cmidrule(lr){8-10}\cmidrule(lr){11-13}
& s/upd. & Total & Mem. & s/upd. & Total & Mem. &
s/upd. & Total & Mem. & s/upd. & Total & Mem. \\
\midrule
DPM         & 2.22 & 31.8 & 48.8 & 1.06 & 14.8 & 18.4 & 0.98 & 21.5 & 11.5 & 0.82 & 17.4 & 9.2 \\
DyDiff      & 2.19 & 31.3 & 48.8 & 1.07 & 15.0 & 18.4 & 0.98 & 21.9 & 11.5 & 0.83 & 17.8 & 9.2 \\
DSSD (Ours) & 2.22 & 31.9 & 48.8 & 1.02 & 14.8 & 18.4 & 1.00 & 22.7 & 11.5 & 0.86 & 17.7 & 9.2 \\
\bottomrule
\end{tabular*}
\end{table}

\subsection{Evaluation metrics}
\label{app:experiments-metrics}

\paragraph{Evaluation protocol.}
The number $K$ of generated futures per input follows the evaluation
protocol of DyDiff \citep{guo2025dydiff}, which adopts the prevailing
convention of each domain. For SEVIR and Turbulence, the $K=8$ generated
futures form a probabilistic ensemble, as in \citet{chen2024cogdpm}. The
finite-ensemble CRPS estimator below accounts for ensemble size, and $K=8$
keeps sampling tractable on the $128\times128$ SEVIR test set. For BAIR and
RoboNet, we follow the video-prediction protocol of
\citet{gupta2022maskvit} and \citet{wu2024ivideogpt}, which selects among
$K=100$ generated videos per input. Because this selection improves with
$K$, we retain $K=100$ for comparability. Within each dataset, all methods
share the same $K$ and 50-step DDIM sampler, so method rankings are
comparable. Scores are not comparable between the two dataset groups.

\paragraph{CRPS.}
For target $y$ and generated futures $x^{(1)},\ldots,x^{(K)}$, the
finite-ensemble probability-weighted-moment CRPS \citep{gneiting2007scoring,zamo2018crps}
at a pixel is
\[
\widehat{\mathrm{CRPS}}(y,x^{(1:K)})
=\frac{1}{K}\sum_{k=1}^{K}|x^{(k)}-y|
-\frac{1}{K(K-1)}\sum_{1\leq i<j\leq K}|x^{(i)}-x^{(j)}|.
\]
Unpooled CRPS averages this score over pixels, channels, and forecast
times, and is reported on all four datasets. On SEVIR and Turbulence, we
also report pooled variants: we pool the target and each generated future
spatially, using either $8\times8$ average or maximum pooling with stride 2,
then compute the same CRPS. The subscripts in $\mathrm{w8}_{\mathrm{avg}}$
and $\mathrm{w8}_{\mathrm{max}}$ identify these two pooling operations. To
retain compatibility with the reference evaluation, the pooled Turbulence
variants include only pooling windows whose target velocities are all
nonnegative; unpooled CRPS uses the full field. The per-horizon BAIR curves
in Figure~\ref{fig:bair-horizons} use frame-wise CRPS after $8\times8$
spatial max pooling, averaged over the full BAIR test set.

\paragraph{CSI.}
CSI is reported on SEVIR and Turbulence. We threshold the fields before
applying a $5\times5$ maximum neighborhood operator with stride 2. SEVIR
uses raw VIL thresholds $16,74,133,160,181,219$; Turbulence uses threshold
zero for each velocity component. Each CSI is
$\mathrm{TP}/(\mathrm{TP}+\mathrm{FP}+\mathrm{FN})$, and the reported score
is macro-averaged over defined threshold--horizon--channel cells; cells
without an observed or predicted event are excluded. We follow the
implemented maximum-neighborhood convention, which differs from an
average-pooling description of CSI in the DyDiff paper \citep{guo2025dydiff}.

\paragraph{Temporal variogram score.}
The temporal variogram score \citep{scheuerer2015variogram} averages, over lags 1--4, the squared
discrepancies between the target's absolute temporal increments raised to
power $1/2$ and the ensemble mean of the corresponding powered absolute
increments. The per-horizon BAIR curves use only consecutive future-frame
pairs and therefore start at the second future frame, because the first has
no preceding future frame.

\paragraph{Energy score.}
Let $\operatorname{dist}(a,b)=\|a-b\|_2/\sqrt{MCHW}$ denote the root-mean-square distance over
a complete $M\times C\times H\times W$ future video. We report the energy score
\citep{gneiting2007scoring}
\[
\mathrm{ES}(y,x^{(1:K)})
=\frac{1}{K}\sum_{k=1}^{K}\operatorname{dist}(x^{(k)},y)
-\frac{1}{2K^2}\sum_{i=1}^{K}\sum_{j=1}^{K}\operatorname{dist}(x^{(i)},x^{(j)}).
\]

\paragraph{RMSE.}
RMSE is defined differently for the two dataset groups. On SEVIR and
Turbulence, it is the RMSE of the ensemble mean of the $K$ generated
futures. On BAIR and RoboNet, it is the square root of the mean per-input
minimum full-video MSE, so a generated future is selected for each input;
the BAIR entries are computed as square roots of the corresponding MSE scores.

\paragraph{PSNR, SSIM, and LPIPS.}
These metrics are reported on BAIR and RoboNet. For each of the $K=100$
generated videos, we average framewise PSNR, SSIM \citep{wang2004ssim}, and VGG-based
LPIPS \citep{zhang2018lpips} over
future time. We then report the maximum PSNR and SSIM and the minimum LPIPS
for each input, selecting independently for each metric, so the selected
video can differ across the three metrics. PSNR and SSIM are evaluated on
the $[0,255]$ RGB scale; LPIPS uses RGB in $[-1,1]$.

\paragraph{FVD.}
Unlike the selection-based metrics above, FVD \citep{unterthiner2018fvd} compares the
distribution of all 100 generated videos per input with the real videos. We prepend the
observed frames to every generated future and to its real counterpart, then
compute the Fr\'echet distance between I3D \citep{carreira2017i3d} feature distributions. LPIPS and
FVD are computed offline from saved predictions. Their implementation
follows the stated aggregation protocol, but exact numerical identity with
an unspecified upstream feature extractor is not assumed.

\subsection{Learned-mode diagnostics}
\label{app:experiments-modes}

Each temporal mode corresponds to one complex-conjugate pole pair, so the six
residue weights index six modes. Figure~\ref{fig:mode-adaptation}(a) reports
the absolute centered residue logits $|\rho_n-\bar\rho|$ and their $L_2$ norm, the
residue-logit norm, on SEVIR. Figure~\ref{fig:residue-pairs} additionally shows the residue shares
$r_n=\operatorname{softmax}(\rho)_n$ for all four datasets. These quantities
are computed from checkpoints saved every 10{,}000 updates from 10{,}000 to
50{,}000 updates. Centering removes the common logit shift that leaves the
shares unchanged. The residue-logit norm measures the spread of the logits around their
mean; absolute centered logits alone do not indicate whether a mode gains
or loses weight. The shares directly show how relative weight is redistributed.

Figure~\ref{fig:mode-adaptation}(b) reports normalized residue entropy,
$-\sum_{n=1}^{6}r_n\log r_n/\log 6$, computed from logged training values
over the same update range. Values near one indicate nearly uniform shares,
while lower values indicate greater concentration. These shares are mixture
weights, not independent mode energies; their entropy does not measure pole
decay or frequency. As a separate check on the near-constant
Turbulence entropy, we reconstruct the discrete pole of its first mode at
fixed $\gamma=0.5$: its radius is $0.937$ at 10{,}000 updates and $0.889$ at
50{,}000 updates, indicating faster decay at that dynamics level despite
little redistribution of residue weight. These parameter changes do not establish physical mode
identification or a causal contribution to forecasting quality.
\section{Additional Results}
\label{app:additional-results}

\subsection{Qualitative forecasts}
\label{app:additional-results-qualitative}

\begin{figure}[p]
\centering
\includegraphics[width=\linewidth,height=0.88\textheight,keepaspectratio]{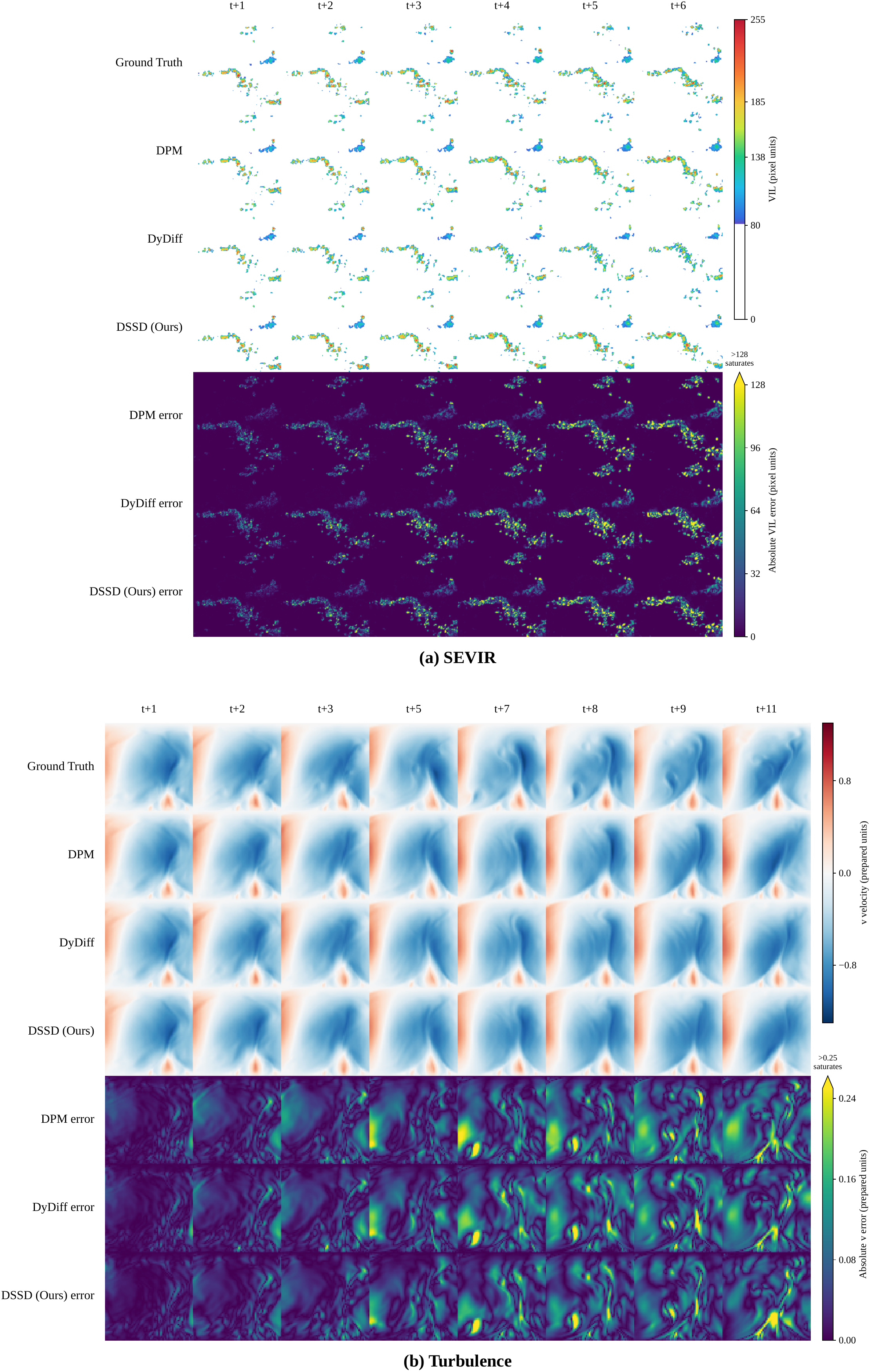}
\caption{Example scientific-field forecasts and absolute errors on (a) SEVIR
and (b) Turbulence. Turbulence shows the $v$ velocity component. Errors above
the color-bar maximum saturate.}
\label{fig:scientific-qualitative}
\end{figure}

\begin{figure}[p]
\centering
\includegraphics[width=\linewidth,height=0.88\textheight,keepaspectratio]{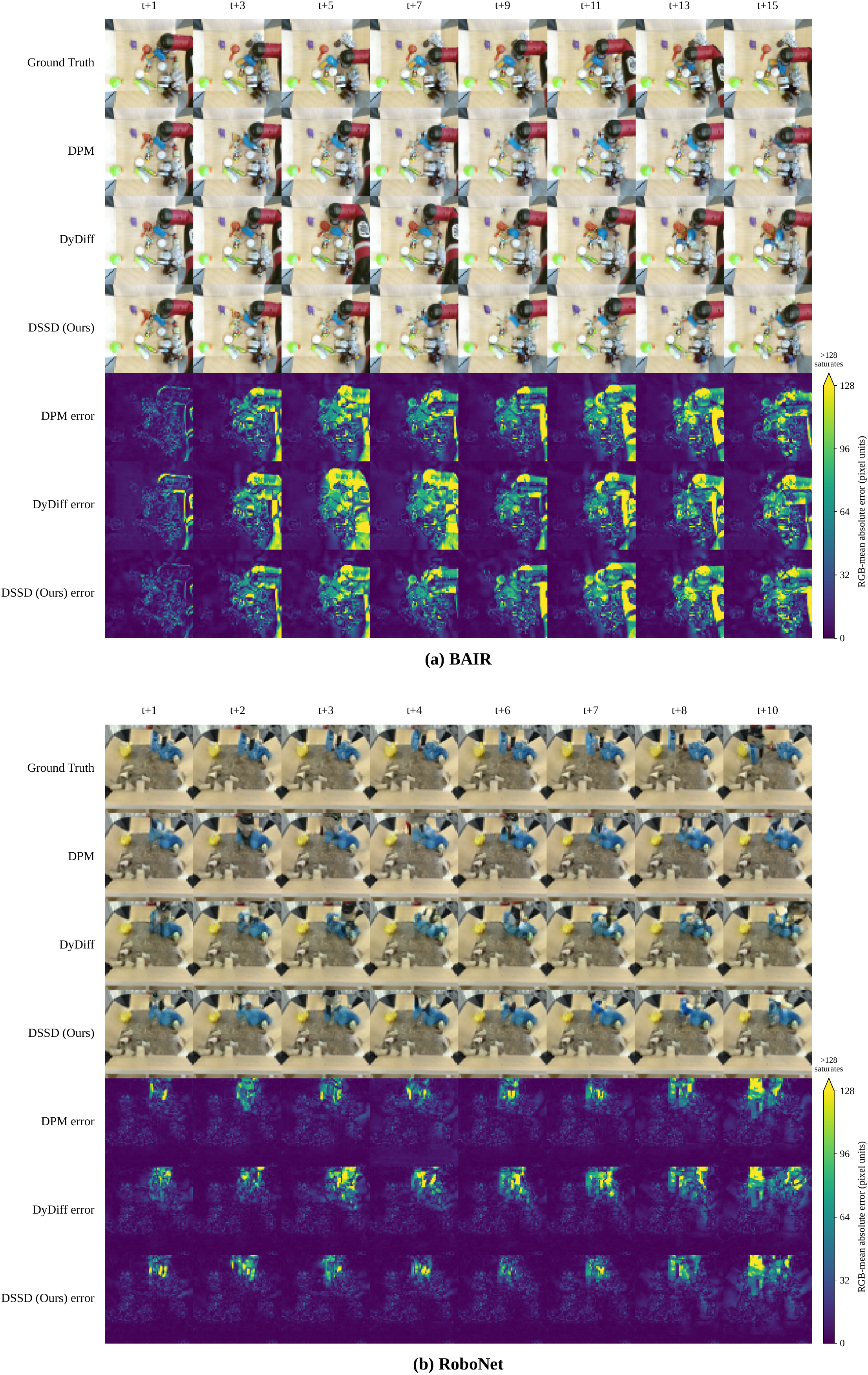}
\caption{Example robot-video forecasts and RGB-mean absolute errors on
(a) BAIR and (b) RoboNet. Errors above the color-bar maximum saturate.}
\label{fig:robot-qualitative}
\end{figure}

Figures~\ref{fig:scientific-qualitative} and~\ref{fig:robot-qualitative}
show one test sequence per dataset. We select the sequence on which DSSD most
improves on the better baseline in per-sequence $8\times8$ average-pooled CRPS.
For SEVIR, we take the highest-ranked sequence whose last two observations and
target each have at most 5\% saturated pixels and a standard deviation of at
least 30 VIL units, because nearly saturated fields are hard to read. Each
method row shows the generated future with the lowest full-video MAE among its
$K$ samples, and the error rows show that sample's absolute error.

In all four examples, errors concentrate in moving structures: storm cells in
SEVIR, shear regions in Turbulence, and the robot arm in BAIR and RoboNet.
The baselines' errors spread more with lead time. This is most visible from
$t{+}5$ to $t{+}9$ in Turbulence, and against DyDiff in the robot videos.
DSSD's errors stay more localized, although they also grow at the final
RoboNet frame.

\subsection{Backbone ablation on SEVIR}
\label{app:additional-results-backbone}

Table~\ref{tab:sevir-backbone} compares the three methods with either a video
U-Net or a Diffusion Transformer (DiT) denoiser \citep{peebles2023dit}, described in
Appendix~\ref{app:experiments-training}.
Within each backbone and metric, bold and underlined values indicate the best
and second-best methods, respectively.

\begin{table}[t]
\centering
\small
\setlength{\tabcolsep}{4pt}
\caption{Backbone ablation on SEVIR. Results with a U-Net and a DiT denoiser. \textbf{Bold}: best; \underline{underline}: second best within each backbone.}
\label{tab:sevir-backbone}
\begin{tabular}{ll ccc c c c c}
\toprule
\multirow{2}{*}{Backbone} & \multirow{2}{*}{Method} &
\multicolumn{3}{c}{CRPS $\downarrow$} &
\multirow{2}{*}{CSI w5 $\uparrow$} &
\multirow{2}{*}{\shortstack{Temporal\\variogram $\downarrow$}} &
\multirow{2}{*}{\shortstack{Energy\\score $\downarrow$}} &
\multirow{2}{*}{RMSE $\downarrow$} \\
\cmidrule(lr){3-5}
& & Raw & w8$_{\text{avg}}$ & w8$_{\text{max}}$ & & & & \\
\midrule
\multirow{3}{*}{U-Net}
  & DPM         & \underline{4.174} & \underline{2.737} & \underline{6.599} & \underline{0.5518} & \underline{2.339} & \underline{10.862} & \underline{15.420} \\
  & DyDiff      & 4.642 & 3.169 & 7.955 & 0.3911 & 4.492 & 13.024 & 25.540 \\
  & DSSD (Ours) & \textbf{3.975} & \textbf{2.421} & \textbf{5.995} & \textbf{0.5707} & \textbf{2.129} & \textbf{10.263} & \textbf{14.585} \\
\midrule
\multirow{3}{*}{DiT}
  & DPM         & 5.425 & 3.862 & 7.453 & 0.4314 & 3.559 & \underline{10.437} & \underline{23.689} \\
  & DyDiff      & \underline{5.260} & \underline{3.789} & \underline{7.327} & \underline{0.4383} & \underline{3.326} & 10.777 & 24.905 \\
  & DSSD (Ours) & \textbf{5.211} & \textbf{3.706} & \textbf{7.188} & \textbf{0.5118} & \textbf{2.784} & \textbf{10.302} & \textbf{23.375} \\
\bottomrule
\end{tabular}
\end{table}

DSSD obtains the best score on all seven reported metrics within both backbone groups. With DiT, it achieves a higher CSI w5 than DyDiff, the strongest baseline on this metric (0.5118 vs. 0.4383), and a lower temporal variogram score (2.784 vs. 3.326). Its raw CRPS is also lower (5.211 vs. 5.260). Thus, the favorable within-backbone ranking observed with the U-Net also holds for the transformer denoiser on SEVIR.

\subsection{Mode-resolved temporal adaptation}
\label{app:additional-results-modes}

\begin{figure}[t]
\centering
\includegraphics[width=0.95\linewidth]{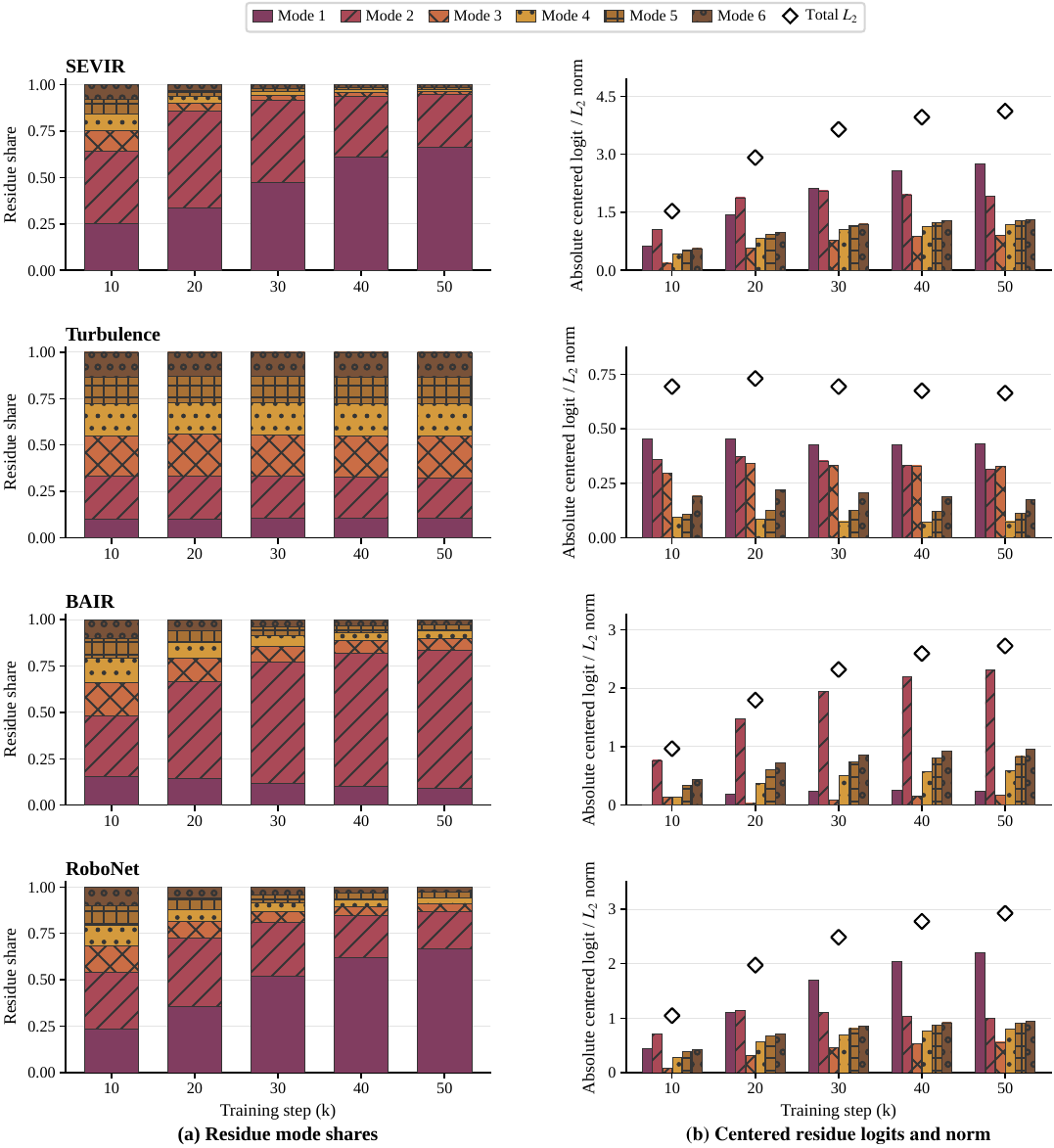}
\caption{Mode-resolved residue diagnostics on the four datasets,
measured every 10k training steps (10k--50k) from saved DSSD checkpoints.
(a) Softmax residue shares of the six temporal modes, each corresponding to a
conjugate pair. (b) Absolute centered residue logits of each mode (bars) and
their total $L_2$ norm (diamonds).}
\label{fig:residue-pairs}
\end{figure}

Each temporal mode corresponds to one complex-conjugate pair of poles, so
the six residue weights index six modes. Figure~\ref{fig:residue-pairs}
tracks how the size and share of each mode change during training, revealing
which modes the model comes to rely on and which it suppresses.
Concretely, it shows the softmax residue shares
$r_n=\operatorname{softmax}(\rho)_n$ at the saved 10{,}000--50{,}000-update
checkpoints, together with $|\rho_n-\bar\rho|$ for each mode and the $L_2$
norm of the centered six-vector. Centering removes the common logit shift,
which leaves the residue shares unchanged.

The trajectories differ across datasets. SEVIR's first-mode share increases
from $0.255$ to $0.664$, BAIR's second-mode share from $0.329$ to $0.744$,
and RoboNet's first-mode share from $0.234$ to $0.667$. Turbulence retains
similar shares across the five checkpoints, and its residue-logit norm
remains near $0.7$. These observations explain the differing entropy trends
in Figure~\ref{fig:mode-adaptation}; they measure residue reweighting, not
independent mode energies. In particular, nearly fixed Turbulence residue
shares do not imply fixed poles, as the pole-radius comparison in
Section~\ref{sec:experiments-modes} illustrates. Mode indices are tracked
within each run and do not identify matching physical modes across datasets.

\subsection{Multivariate time-series forecasting}
\label{app:additional-results-timeseries}

We evaluate multivariate forecasting on six datasets distributed through GluonTS \citep{alexandrov2020gluonts}, following TimeGrad's dataset and prediction length setting \citep{salinas2019copula,rasul2021timegrad}. The datasets are Exchange, Solar, Electricity, Traffic, Taxi, and Wikipedia. To hold the representation and denoiser fixed across methods, DPM, DyDiff, and DSSD use the same temporal 1D dilated-convolution network: a pointwise channel projection followed by eight residual blocks with 128 hidden channels, which jointly processes observed values and noisy future values.

Each dataset is one multivariate sequence with $D$ channels. We compute a mean and standard deviation for each channel from the training sequence and apply the same normalization to test data. For each GluonTS test window, we forecast its future $M$ points from the preceding $N$ points. Table~\ref{tab:ts-50k} gives these dimensions and window lengths.

All methods are trained from scratch for 50{,}000 optimizer updates with batch size 16, Adam at a constant learning rate of $10^{-5}$, a 1{,}000-step cosine diffusion schedule, and seed 42. DyDiff uses the fixed EMA operator with $\eta=0.5$. For DSSD, we use three conjugate SSM mode pairs (state size six), compared with six pairs in the video experiments. The time-series representation has no spatial latent feature maps, so we use fewer temporal modes as a modest capacity choice. All methods share the same denoiser. We use 50-step deterministic DDIM with a linearly spaced timestep grid.

Table~\ref{tab:ts-50k} reports channel- and time-averaged finite-ensemble PWM CRPS, ensemble-mean RMSE and MAE, energy score, and temporal variogram score. Metric definitions follow Appendix~\ref{app:experiments-metrics}.

\begin{table*}[t]
\centering
\small
\setlength{\tabcolsep}{5pt}
\caption{Multivariate time-series test results at 50{,}000 updates.
Bold and underlined values are best and second best within each dataset.
All scores use training-set normalization; lower is better. $D$ is the
number of jointly modeled series; $N$ and $M$ are observed and predicted
time steps.}
\label{tab:ts-50k}
\resizebox{\textwidth}{!}{%
\begin{tabular}{lrrrlrrrrr}
\toprule
Dataset & $D$ & $N$ & $M$ & Method & CRPS $\downarrow$ & RMSE $\downarrow$ & MAE $\downarrow$ & Energy $\downarrow$ & Variogram $\downarrow$ \\
\midrule
\multirow{3}{*}{Exchange} & \multirow{3}{*}{8} & \multirow{3}{*}{30} & \multirow{3}{*}{30} & DPM & \underline{0.37573} & \underline{0.68591} & \underline{0.56123} & 0.59404 & 0.78559 \\
 & & & & DyDiff & 0.37640 & 0.76033 & 0.64038 & \underline{0.54593} & \underline{0.60261} \\
 & & & & DSSD (Ours) & \textbf{0.29896} & \textbf{0.61315} & \textbf{0.51080} & \textbf{0.46610} & \textbf{0.44125} \\
\midrule
\multirow{3}{*}{Solar} & \multirow{3}{*}{137} & \multirow{3}{*}{24} & \multirow{3}{*}{24} & DPM & \underline{1.43163} & \underline{2.31897} & \textbf{1.95040} & \underline{1.89547} & \textbf{2.25310} \\
 & & & & DyDiff & 1.48259 & 2.34964 & 1.97462 & 1.92896 & 2.37242 \\
 & & & & DSSD (Ours) & \textbf{1.41653} & \textbf{2.31678} & \underline{1.95240} & \textbf{1.88563} & \underline{2.26805} \\
\midrule
\multirow{3}{*}{Electricity} & \multirow{3}{*}{370} & \multirow{3}{*}{24} & \multirow{3}{*}{24} & DPM & \underline{7.12656} & \underline{7.08183} & \underline{5.72460} & \underline{6.72375} & \textbf{8.66073} \\
 & & & & DyDiff & 7.17127 & 7.10774 & 5.74584 & 6.74831 & \underline{8.67150} \\
 & & & & DSSD (Ours) & \textbf{7.04436} & \textbf{7.05699} & \textbf{5.70272} & \textbf{6.69563} & 8.87129 \\
\midrule
\multirow{3}{*}{Traffic} & \multirow{3}{*}{963} & \multirow{3}{*}{24} & \multirow{3}{*}{24} & DPM & \underline{7.12199} & \underline{10.90963} & 9.67041 & \underline{8.47188} & \underline{6.88153} \\
 & & & & DyDiff & 7.14840 & 10.91133 & \textbf{9.66698} & 8.47718 & \textbf{6.85649} \\
 & & & & DSSD (Ours) & \textbf{7.08377} & \textbf{10.90212} & \underline{9.66832} & \textbf{8.46117} & 7.08459 \\
\midrule
\multirow{3}{*}{Taxi} & \multirow{3}{*}{1{,}214} & \multirow{3}{*}{24} & \multirow{3}{*}{24} & DPM & \underline{4.40064} & \underline{5.75505} & 4.91774 & \underline{4.65397} & \underline{3.57764} \\
 & & & & DyDiff & 4.41067 & 5.75627 & \underline{4.91771} & 4.65638 & \textbf{3.57039} \\
 & & & & DSSD (Ours) & \textbf{4.37568} & \textbf{5.75023} & \textbf{4.91415} & \textbf{4.64667} & 3.68216 \\
\midrule
\multirow{3}{*}{Wikipedia} & \multirow{3}{*}{9{,}535} & \multirow{3}{*}{30} & \multirow{3}{*}{30} & DPM & \textbf{10.35604} & 16.95278 & 11.20462 & 12.66044 & \textbf{9.40709} \\
 & & & & DyDiff & 10.41862 & \textbf{16.70496} & \textbf{11.03162} & \textbf{12.54705} & \underline{9.47266} \\
 & & & & DSSD (Ours) & \underline{10.41440} & \underline{16.93813} & \underline{11.19473} & \underline{12.65975} & 9.52240 \\
\bottomrule
\end{tabular}
}
\end{table*}

At 50{,}000 updates, DSSD has the lowest CRPS, RMSE, and energy score on five of six datasets, with DPM leading CRPS on Wikipedia and DyDiff leading its RMSE and energy score. The temporal variogram score is mixed: DSSD is best on Exchange and second on Solar, but has the highest score of the three methods on Electricity, Traffic, Taxi, and Wikipedia.

\end{document}